\pdfoutput=1

\documentclass[11pt]{article}

\usepackage[utf8]{inputenc}
\usepackage[T1]{fontenc}

\usepackage[margin=1in]{geometry}
\usepackage{setspace}
\usepackage{microtype}

\usepackage[colorlinks=true,citecolor=blue,linkcolor=blue,urlcolor=blue]{hyperref}
\hypersetup{pdftitle={The Announcement Carries the Cue: Markup, Boundaries,
    and the Notation of Pre-Training Corpora},
  pdfauthor={E. M. Freeburg}}
\usepackage{url}
\usepackage[numbers]{natbib}

\usepackage{booktabs}
\usepackage{tabularx}

\usepackage{graphicx}
\graphicspath{{figures/}{./}}   

\usepackage{listings}
\usepackage{xcolor}

\usepackage{enumitem}

\usepackage{amsmath}

\definecolor{wavered}{rgb}{0.80,0.25,0.05}
\definecolor{provblue}{rgb}{0.10,0.30,0.55}

\newcommand{\Sw}[1]{$S(#1)$}
\newcommand{\ci}[2]{\,[$#1$,\,$#2$]}

\title{\textbf{The Announcement Carries the Cue}\\[0.35em]
  \large Markup, Boundaries, and the Notation of Pre-Training Corpora}

\author{
  E. M. Freeburg \\
  Independent Researcher \\
  \texttt{hqops@icloud.com}
}

\date{August 2026}

\begin{document}

\maketitle

\begin{abstract}
How a document's arrangement is written down, its \emph{notation}, is a
training variable that no dataset card records. The field has established
that text-extraction choices change model behaviour, and has never once
measured the notation of what those choices put into the corpus. We define clean-window
survival, a deterministic count of how much of a stream still demands the
boundary inference, and measure notation on three fronts. \emph{What corpora
carry:} a census of thirteen public corpora, where survival falls to 0.153
in a vision-converted PDF slice against 0.889 in C4; the scarce
resource is not unmarked text but long unmarked text; a pre-registered
supply test finds what remains institutional, not consumer. Our own
pre-registered prediction
failed: converters do not fabricate structure on prose, and that null forced
the reliability mechanism that survives it. \emph{What readers use:} across
five base models spanning 0.6B to 8.2B and two pipelines, deleting a
structural announcement makes the following prose measurably harder to
predict, while swapping its notation moves nothing. That zero does not make
notation unimportant; it relocates the variable: the operative cue is the
announcement, not the sigil. \emph{What writers impose:} a bounded null.
Base models do not impose the marked register above the authored baseline,
and handed prose with every announcement deleted they do not put one back,
at a rate indistinguishable from zero against an authored reference of
zero. We ship the format those measurements imply: the pure frame,
paragraphs in authored order, every announcement deleted into a reversible
sidecar, mixed against the marked copy over announcement presence rather
than notation. Choose format operators by the capability they train,
not by the fidelity they preserve, and record extractor identity and
survival on data cards.
\end{abstract}

\section{First paragraph to last}
\label{sec:crux}

Take \emph{Moby-Dick} from its first paragraph to its last and delete
everything that is not the book: the chapter numbers, the chapter titles, the
running heads, the transcriber's furniture. On the Project Gutenberg text
that is 143 lines (141 chapter heads and two thematic breaks) and 1,442
tokens. What remains is 283,267 tokens of
Melville in authored order, and the deletion is 0.51\% of the
file.\footnote{This paper allows itself one em dash per paragraph and no
more. The prior paper made the same promise, for the same reason: the em dash
is the most legible residue of the output notation this work is about, and a
paper about notation should be legible about its own \citep{freeburg_2026}.}
A reader handed the result would not notice that anything had happened to it.
A model handed the result is handed something it has almost never seen: a
continuous stretch of 283,267 tokens in which nothing on the surface
announces where one part of the book ends and the next begins.

That ratio is the argument of this paper stated once, and what the deletion
does is not subtle. Packed at the 8,192-token window most pre-training still
uses, the book arrives as 35 fragments carrying 4.1 announcements each, and
every one of its 34 full windows contains at least one. After the deletion
all 34 contain none, clean-window survival against ground truth goes from
0.000 to 1.000 at both 8,192 and 32,768 tokens, and the longest span
containing no announcement goes from 10,595 tokens to the entire book.

The middle case is the one that reorganized this paper. Strip the Markdown
but leave \texttt{CHAPTER 1. Loomings.} standing on its own line, and the
census statistic the field would report for that file reads 1.000: perfectly
clean, indistinguishable from the deleted version. Against ground truth the
same file reads 0.000, exactly the marked book, because every boundary is
still there, in the same place, on its own line. Faithful flattening moves
the notation and leaves the announcement, and the announcement is what the
reader was using.

The deletion is not free, and its price should be quoted before its benefit.
Of those 1,442 tokens, 695 are label apparatus carrying no content and 747
are Melville's own title words. Every deleted line goes into a sidecar, the
transform is reversible, and in the mixture this paper recommends the marked
copy coexists with the pure one, so the corpus loses nothing and only the
pure \emph{copy} does. How big the deletion is depends on how a book
announces itself: across the 27-work pack of \S\ref{sec:pureframe} it is
0.71\% of tokens, 0.25\% across the fifteen narratives, and 2.6\% in the
worst single case; had every work announced itself with bare labels rather
than titles, the whole operator would have cost 0.048\% of the pack.
It is under one percent almost everywhere, and the size is never the point.
What is left is the point.

It is also, we think, the book as a person actually holds it. Nobody
remembers the fifth chapter of \emph{Moby-Dick}; they remember a part of the
story in the context of the whole story. Readers plainly do see the breaks,
and we are not claiming otherwise. The claim is about what comprehension and
memory run on, and the reading literature is consistent in treating
segmentation as something a reader constructs continuously out of the
content rather than something the page supplies
\citep{zacks_2007,zwaan_radvansky_1998}. We take that as motivation and not
as evidence.\footnote{Chapter grids are frequently production artifacts.
Nineteenth-century serial publication set chapter length to the size of an
instalment, which is the same species of decision a conversion pipeline
makes today, made for the same species of reason.}

The reason nobody has run the corresponding experiment is not that it is
expensive. It is that the variable is not in the record. The field has
established that text-extraction choices change model behaviour, and has
never once measured the notation of what those choices put into the corpus.
No dataset card
for C4, The Pile, RedPajama, Dolma, FineWeb, DCLM, Nemotron-CC, FinePDFs or
the olmOCR corpora reports how much structural markup its text carries, and
no statistic of that family appears anywhere in the literature. The most
ambitious corpus audit of 2026 makes the point better than an assertion
could: propella-1 annotates three billion documents across FineWeb-2,
FinePDFs, HPLT 3.0 and Nemotron-CC on eighteen properties in six categories,
and not one of the six categories is notational \citep{propella_2026}.

The neighbouring literature runs the other way, and we concede it completely
before doing anything else. Structure-preserving extraction makes models
better on every axis anyone has measured. A model-based HTML parser is worth
1.08 percentage points over trafilatura-extracted text across thirteen
benchmarks \citep{aicc_2025}; extractor choice alone is worth up to ten
percentage points on WikiTQ \citep{beyond_single_extractor_2026}. We do not
contest a single one of those results. We change the object and the
variable. Those studies measure the consequence of an extraction choice on a
benchmark, and nobody has measured the notation of the corpus that the
choice produces;
their variable is markup \emph{richness}, which is a fidelity property,
while ours is markup \emph{reliability}, which is a distributional one. AICC
is the result our mechanism explains rather than the rival it beats.
Preserving markup helps on every task that reads structure, and the question
here is what it costs at the one stage of training where structure has to be
inferred.

This paper measures the variable on three fronts and then ships the format
the measurements imply. Our own prior asserted the corpus thesis from the
model side without measuring it \citep{freeburg_2026}; this is the
measurement. Four things are new.

\begin{itemize}[leftmargin=1.4em,itemsep=2pt,topsep=3pt]
\item \textbf{An instrument and a census.} Clean-window survival $S(W)$ and
  the clean-run distribution: a deterministic surface count, no model, no
  annotation, no seed, over thirteen public corpora. Survival falls to
  0.153 in the olmOCR slice of Dolma~3, whose documents carry markup at
  74.0\%, against 0.889 in C4. Anyone can recompute it in an afternoon and get the same
  number, which is why the documentation ask at the end is realistic.
\item \textbf{Three fronts on one variable}, honest nulls led: what corpora
  carry (\S\ref{sec:front-corpora}), what readers use
  (\S\ref{sec:front-readers}), what writers impose
  (\S\ref{sec:front-writers}). Between them sits a pre-registered converter
  study whose prediction was ours and failed, and a notation swap that moves
  nothing because the operative cue is the announcement rather than the
  sigil.
\item \textbf{A format, specified, validated and shipped.} The pure frame:
  paragraphs in authored order, every structural announcement deleted, every
  deletion recorded in a reversible sidecar, and a validator that passes a
  file only if it contains no announcements at all.
\item \textbf{A data card for notation}, which is mostly a request to finish
  a field that the field's own flagship PDF corpus already carries.
\end{itemize}

\section{The variable}
\label{sec:variable}

A training corpus is not a collection of documents. It is a collection of
images of documents under format operators: extraction, conversion,
flattening, OCR, transcription convention, synthetic rephrasing. Write a
document as a pair, a content $c$ and the operator $f$ that produced the
version stored on disk. The training distribution is over the images $f(c)$
alone. Domain labels are metadata about $c$, an object the gradient never
touches, so two corpora with identical domain marginals and different
operator distributions are different training distributions, and nothing on
a standard dataset card would show it.\footnote{Operators are recorded in the
tokens whether or not anyone records them anywhere else. The 1851 printer of
\emph{Moby-Dick} set em dashes. The transcriber who prepared the PG-19
edition turned every one into a double hyphen, 1,714 of them, leaving no em
dash in the book \citep{pg19_2019}; the transcriber who prepared the copy we
downloaded turned them back, 1,730 em dashes and no double hyphens. Our own
marked serializer then writes two \texttt{---} thematic breaks into the same
work. One book, four token records, and not one of the four operators is
recorded where a downstream consumer could read it.}

Call the missing quantity \emph{notation}: the distribution, across a
corpus, of how documents' arrangement is written down. It is not a
proposal for tidier bookkeeping. It names a decision that is currently
made by accident, at no apparent cost, by people optimizing something
else.

Notation matters here for one specific reason. A Markdown heading is a
two-token cue that is present at every discourse boundary and, in the
material this paper is about, close to absent everywhere else. It is
trivially detectable, faithful converters place it where the author put a
boundary, and conversion pipelines now attach it to essentially everything
long enough to have structure. Consider what a model has to do with a chapter
break when nothing announces it: the prose cools, a thread resolves, a new
scene opens with a widened aperture, and the model, to predict well, must
represent that a boundary of a certain kind has occurred, because everything
downstream depends on it. That representation has to be assembled from
evidence spread over thousands of tokens. It is, in the plainest sense, an
inference. Put \texttt{\#\#} in front of the chapter title and the inference
is no longer required. How reliable that cue actually is turns out to be
measurable, and \S\ref{sec:invariance} measures it: it holds in narrative and
comes apart elsewhere.

The shortcut-learning literature is explicit that what governs whether a
model takes a feature is not how often the feature appears but how reliably
it predicts and how easily it can be extracted
\citep{geirhos_2020,hermann_2024}, and the induction-head scaling work puts
a number on the same trade: circuit formation is governed jointly by
repetition frequency and repetition reliability, with sensitivity to
reliability more than twice the sensitivity to frequency
\citep{aoyama_2026}.\footnote{We cite version three of
\texttt{arXiv:2511.16893} (ICML 2026), Eq.~7 and \S7; the v1 posting, which
\texttt{ar5iv} still renders, gives different exponents.} A structural sigil
(the markup character itself: the \texttt{\#\#} before a chapter title, the
bullet, the horizontal rule) is the limit case on both axes: maximally
available, being a literal surface token, and near-perfectly predictive,
because faithful converters place it at every real
boundary and, in narrative, almost nowhere else; \S\ref{sec:invariance} puts
that precision at 0.967 once the documented typographic furniture is removed
from the ground truth. Reliability, not volume, is what a shortcut runs on.
That is
the whole reason a variable worth 0.065\% of the tokens pooled across the
fifteen serialized narrative works of \S\ref{sec:nonull} can be worth
measuring.

Two properties make the situation hard to reach from where the field
stands. The first is that it is \textbf{unreachable}. Every method for
controlling what a model learns from its data operates over documents as
they already sit in the corpus: mixture weights decide how often a document
is sampled, filters decide which are kept, and both take the corpus as
given. The operation at issue happened before that.

\begin{quote}
Reweighting operates over documents as built, and therefore cannot reach
upstream of the operator that built them. Only re-extraction can, and
nobody re-extracts.
\end{quote}

One operation does survive downstream, and \S\ref{sec:pureframe} is built on
it: deletion. An announcement still present in the stream can be removed and
recorded; notation already destroyed cannot be recovered without the source.

The second is that it is \textbf{unpriced}. The title stays in the corpus,
the domain weight is unchanged, the token count is unchanged, and no
dataset card records what a document passed through on the way in. Every
benchmark in current use rewards the arrangement that structure-preserving
conversion produces, so published measurements report structural
normalization as an improvement, and under the instruments in use they are
right. A field hill-climbing on a proxy that is blind to one side of a
trade keeps climbing, and every step is correctly reported as progress. The
burden that places on this paper is worth stating plainly: if no instrument
can price the other side, the claim is unfalsifiable and should be
discarded. Section~\ref{sec:instrument} proposes the instrument.

One distinction has to be fixed before any of this can be measured, because
the field currently collapses it. \emph{Serialization register} is the
notation a text is stored in, and pipelines set it. \emph{Authorial
composition register} is whether the writer was thinking in structure at all,
and history sets it. The two come apart at both ends: Books3 shipped
novelists' prose wearing the pipeline's headings \citep{presser_epub2txt},
and PDF erases the authoring environment by construction \citep{iso32000},
after which extractors converge everything onto a common notation, so a
\LaTeX{} paper and a word-processor brochure arrive marked up the same way.
Pipelines cannot create authorial composition register; they can only
preserve or destroy the evidence of it, which is why this has to be recorded
at ingestion or not at all. Because the bare word \emph{register} already
has an occupant in this literature \citep{register_matters_2025}, we use it
alone only in that occupant's sense, for the text-type strata of the
ground-truth corpus, and otherwise qualify it in full, except where a
pre-registered rule name fixes its own wording.

Two of our own measurements then relocate the target, and they are the
reason this paper ends where it does. Flattening a novel's Markdown while
keeping its chapter titles as bare lines changes clean-window survival by
essentially nothing on prose (\S\ref{sec:invariance}), and swapping the
notation on identical text changes a trained reader's use of long-range
context by a measured zero (\S\ref{sec:front-readers}). The sigil is not
the cue. The \emph{announcement} is: a short line, set off by whitespace,
that says a boundary is here. Whatever is done about this variable has to
be done at the level of announcements, and that is the level at which
\S\ref{sec:pureframe} specifies a format. Until that format is built, we
give the thing the corpora lack a name: the \emph{missing format}, long text
in which nothing announces the arrangement, so the boundary inference is
demanded rather than optional.

\section{Nobody decided this}
\label{sec:genealogy}

The flatness of the old web corpora was an accident of tooling. C4 consumed
Common Crawl's pre-extracted plain text and then applied a line filter:
``We only retained lines that ended in a terminal punctuation mark''
\citep{t5_c4_2020}. That discards every heading, because headings do not end
in periods. It was a quality heuristic, and it was the largest single
destroyer of structural markup in the corpus that trained a generation of
models. The Pile's authors named the problem in 2020, calling the same
filter ``too aggressive'' in their own appendix \citep{pile_2020}, and
nobody has used the observation since.

The choices that followed were made the same way, on yield. The Pile
extracted Common Crawl with jusText, chosen ``based on visual inspection''
of the output \citep{pile_2020}. FineWeb re-extracted from raw HTML with
trafilatura, through a wrapper that passes no output-format argument and so
leaves the library in its default plain-text mode \citep{datatrove_2026,
fineweb_2024}. That default is not symmetric: trafilatura's serializer
applies heading prefixes only when formatted output is requested, while list
markers are emitted unconditionally \citep{trafilatura_2026}. DCLM chose
resiliparse because it was eight times faster than trafilatura at the same
benchmark quality, and its own appendix records that resiliparse ``keeps in
about 10\% more text, some of which may provide useful content such as
section titles'' \citep{dclm_2024}. Nemotron-CC chose jusText because it
yielded 28.6\% more high-quality tokens \citep{nemotroncc_2025}, and its
mathematics successor changed extractor again, specifically to stop
destroying notation that mattered \citep{nemotroncc_math_2025}. Every one of
these decisions improved the number being watched. Not one was recorded as a
decision about how much structural notation the training stream would carry.

One decision in the record was deliberate, and it is the strongest
counter-current to any story of accidental flattening. The Llama 3 pipeline
removed Markdown outright: markdown is ``harmful to the performance of a
model that is primarily trained on web data compared to plain text, so we
remove all markdown markers'' \citep{llama3_2024}. In the same paragraph,
HTML pages with mathematics and code are processed to ``preserve the
structure'' of that content. A frontier lab stripped notation where the
structure was not authorially real and preserved it where it was, on
benchmark evidence, which is this paper's preserve-and-impose distinction
performed without the variable being named.

Then the direction reversed. The frontier of corpus construction moved from
stripped web text to PDF conversion by vision-language models, and those
models emit structure by design, because faithfulness to the document is the
point \citep{olmocr_2025,docling_2024,mineru_2024}. FinePDFs runs Docling
for text-embedded PDFs and RolmOCR, a Qwen2.5-VL fine-tune on olmOCR's own
training mixture, for the scanned path \citep{finepdfs_2025,rolmocr_2025}.
Qwen3's own account describes fine-tuning Qwen2.5-VL to extract text from
PDFs at pre-training scale \citep{qwen3_2025}. The reversal coincided with a
withdrawal: books were roughly fifteen percent of The Pile in 2020 and 4.5\%
of the LLaMA-1 mixture in 2023 \citep{pile_2020,llama1_2023}, then 6.0B of
Dolma's 3,059B tokens in 2024, which is 0.196\% \citep{dolma_2024}, and
Dolma~3 has no books source at all. The one component that reliably supplied
book-length unmarked prose left the mixture in the same generation that the
most heavily marked material anyone has built entered it.

The mechanism is easiest to see in four lines of a public script. Books3,
which is 12.07\% of The Pile and the largest book source in the most-cited
open corpus of its era, was produced by \texttt{epub2txt-all}
\citep{presser_epub2txt}:

\begin{lstlisting}[language=Python]
h = html2text.HTML2Text()
h.body_width = 0
text = h.handle(html)
bookname = filename + '.md'
\end{lstlisting}

\texttt{html2text} is the canonical HTML-to-Markdown converter, running here
at library defaults, which means \texttt{\#} and \texttt{\#\#} headings,
\texttt{*} bullets, \texttt{**bold**} and \texttt{[text](url)} links
\citep{html2text}. Every chapter file it wrote was named \texttt{.md}. A
default output mode, chosen by nobody, decided the notation of the largest
book corpus in open pre-training, and the script's own header comment traces
the same converter back to BookCorpus.

\section{What to measure}
\label{sec:instrument}

The variable the record lacks is easy to define and cheap to compute. For a
tokenized stream, let clean-window survival $S(W)$ be the fraction of
non-overlapping $W$-token windows containing no structural markup: no
heading sigils, no list markers, no rules, no blockquote prefixes. Alongside
it, measure the distribution of \emph{clean runs}, the maximal spans of
consecutive tokens with no markup at all. $S(W)$ asks how much of a stream
still demands the boundary inference at horizon $W$; the clean-run tail asks
how long the demand is ever sustained. Code fences and table pipes are
tracked separately and excluded from the core tag set, because they mark
content rather than arrangement. Every number in this paper uses the
\texttt{o200k\_base} tokenizer.

One consequence of that definition has to be stated, because it changes what
the statistic means for short-document corpora: a document shorter than one
window contributes no windows at all, rather than counting as trivially
clean.\footnote{\texttt{src/cfc/contamination.py::survival} returns
\texttt{(0,~0)} when \texttt{n\_tokens // W} is zero. The alternative, scoring
a 244-token document as a clean 8,192-token window, would make survival rise
with fragmentation, which is the opposite of what it is for.} For a web
corpus whose median document is 244 tokens, $S(8{,}192)$ is therefore a
statement about the long tail of that corpus and not about its typical
document, while for a books corpus whose median document is 77,297 tokens it
is a statement about the typical document. We report it that way throughout,
and the data-card field of \S\ref{sec:datacard} asks for the
contributing-document share alongside the value for exactly this reason.

The same two statistics can be computed against notation or against
typography, and the difference between those two readings is a measurement
in its own right (\S\ref{sec:invariance}). Against notation, a window is
contaminated if it contains Markdown syntax. Against typography, it is
contaminated if it contains a line that a notation-blind detector reads as
structural: a short line, standing between blank lines, that does not end
the way running prose ends.\footnote{Detector specification, as
implemented. A candidate is a line between blank lines, at most 100
characters, that does not end in prose-terminal punctuation. The terminal
set is the period, exclamation mark and question mark of C4's filter, plus
the comma, colon, semicolon and ellipsis, plus the hyphen, en dash and em
dash, plus the closing quotation mark in both its straight and its curly
form, the closing apostrophe, the parenthesis and the bracket.
The period carries one exception: a period preceded by an uppercase letter
is a label period, as in \texttt{CHAPTER IV.}, and does not disqualify the
line. Any line already matching Markdown syntax also counts. The detector is
therefore close to the complement of C4's line filter, which is not a
coincidence: the filter and the detector are looking for the same lines, for
opposite reasons.}

The detector's precision and its recall are both measurable, and one of them
is a warning about every survival figure in this paper. On a novel whose
chapters are titled rather than numbered, it catches 4 of 141 real
announcements, or 2.8\%, because \texttt{CHAPTER 1. Loomings.} ends in a
lowercase word and a period and therefore reads to the filter as a sentence;
its label-period exception only fires after a capital, as in
\texttt{CHAPTER IV.} Titled heads are largely invisible to it. Every
structural survival figure reported here for material with titled chapters is
therefore an \emph{upper bound} on cleanliness, and the true contamination is
worse than the number. We use the detector as a screen, report its recall
where we know it, and use ground truth wherever ground truth exists.
Rule-class recognition of chapter heads on this corpus family has published
precedent: Chapter Captor segments 9,126 Project Gutenberg novels with
hybrid rule-plus-neural header detection at F1 0.77
\citep{chaptercaptor_2020}, so a deterministic detector is here the field's
own instrument rather than an eccentric one.

The obvious objection is that $S(W)$ is a regular expression, and the answer
is that this is the point. It is deterministic, seedless and auditable;
anyone can recompute it on any corpus in an afternoon and get the same
number to the last digit. The methodological sibling to this work profiles
three million passages of Dolma on eleven discourse dimensions using a
trained scorer built from four hundred hand-annotated passages
\citep{narradolma_2026}, which is a real advance and answers a different
kind of question. It also cannot be asked of a lab: you cannot ask a
pre-training team to run your RoBERTa over their corpus and reproduce your
annotation guidelines. You can ask them to report a surface count over
tokens their pipeline already has in memory, which is why the documentation
proposal in \S\ref{sec:datacard} is realistic and the alternative is not.

\section{Front I: what corpora carry}
\label{sec:front-corpora}

\subsection{The census}
\label{sec:census}

Table~\ref{tab:census} is the census. Ten rows are the published measurement;
three more were run for this paper and are reported separately, at a
different budget and sampling rule, because they are not part of the
published ten. Three facts organize it.

\textbf{The direction of travel is steep.} Between FineWeb and the olmOCR
slice of Dolma~3, the modern web baseline and the vision-converted frontier,
ATX heading presence goes from 0.12\% of documents to 33.3\%, and any-markup
presence from 21.5\% to 74.0\%. On the re-checked FineWeb figure
of 0.191\% the heading swing is a factor of 174, so what the table supports
is a swing well over a hundredfold rather than the cleaner-sounding number
the published rates alone would give. The gradient tracks the pipeline
history exactly: line-filtered plain text at one end, vision-model PDF
conversion at the other.

\textbf{The scarce resource is not unmarked text; it is \emph{long} unmarked
text.} Every web corpus, at every markup level, has a 99th-percentile clean
run between about 2,400 and 6,500 tokens, because a corpus of 244-token
documents cannot contain a long clean run however it is notated. Books are
different in kind: a median document of 77,297 tokens and a 99th-percentile
clean run of 109,045, which is about seventeen times the best web corpus and
roughly twenty times the typical one. The gap is dominated by document length
rather than by markup: no web corpus in the table has a median document
within two orders of magnitude of the books row, so a long clean run is not
available to be had however the web text is notated.

\textbf{No source supplies all three properties at once.} The web is
unmarked and short. The olmOCR slice is long and carries the lowest survival
of anything measured here. Books are long, coherent and low-density, and
books are the row that left the mixture.

The provenance of those ten rows has to be stated, and stating it costs less
than being caught. The original census run wrote no per-row output file.
Seven of the ten rows therefore rest on a measurement that cannot be
regenerated: the sample size, the document count, the sampling rule, the
corpus revision and the date were never recorded, and the only
machine-readable copy of the table before this paper was an array of
constants inside a figure script. That is exactly the failure the paper's
own recommendation exists to prevent, and a paper asking laboratories to
report $S(W)$ on their data cards cannot ship $S(W)$ without an artifact.
Every row now carries its provenance explicitly in
\texttt{data/census/census-table.json}, including the nulls where nothing was
recorded, and six of the ten have been independently re-measured.

Table~\ref{tab:recheck} reports those re-measurements, including the one that
disagrees. The three web rows agree on every quantity, with two caveats we
would rather print than have found: at an 8M-token sample only documents of
at least 8,192 tokens contribute windows, so the survival column rests on 23,
62 and 138 windows for C4, FineWeb and DCLM, and the agreement there is
nearly uninformative; and FineWeb's heading rate re-checks 1.6 times higher
than the published value, which is the correction that moves the swing above.
The olmOCR row diverges. A topic-balanced re-measurement puts its
$S(8{,}192)$ at 0.271 rather than 0.153, and a 25M-token pass on 2,180
windows puts it at 0.296; its any-markup rate re-checks within five points,
and its median document at roughly half. We report the divergence as the
pre-stated criteria classify it. The claim the row is doing work for survives
it: at 0.271 the olmOCR slice is still the lowest-survival corpus in the
census, at half the next-lowest re-measured row, DCLM at 0.558, and it is
still the corpus that fills the one training stage where a document is
experienced whole (\S\ref{sec:slot}).

One note about compositional practice is worth carrying forward. Within
long-form prose, sustained argument supplies longer clean spans than
narrative does, because a treatise's sections run longer than a novel's
chapters. In the ground-truth corpus below, unbroken paragraph runs have a
median of 25 and a 90th percentile of 78 in narrative against a median of 2
in technical writing. The kind of writing that best supplies the missing
format is also the kind that has most completely vanished from modern
mixtures.

\begin{table}[t]
\centering
\small
\begin{tabular}{llrrrrr}
\toprule
corpus & class & median doc & any markup & ATX \texttt{\#} & \Sw{8\mathrm{k}} & p99 clean run \\
\midrule
C4$^{\dagger}$              & web   & 244    & 4.0\%  & 0.02\%  & 0.889 & 3,232   \\
Dolma CC                    & web   & 426    & 2.9\%  & 0.02\%  & 0.823 & 5,790   \\
RedPajama v2                & web   & 555    & 5.7\%  & 0.08\%  & 0.706 & 6,548   \\
Nemotron-CC                 & web   & 328    & 10.7\% & 0.07\%  & 0.568 & 5,025   \\
FineWeb$^{\dagger}$         & web   & 388    & 21.5\% & 0.12\%  & 0.603 & 2,374   \\
DCLM$^{\dagger}$            & web   & 614    & 27.2\% & 0.98\%  & 0.505 & 4,211   \\
The Pile                    & web   & 390    & 15.1\% & 3.97\%  & 0.301 & 5,326   \\
\textbf{Dolma books}$^{\dagger}$ & books & \textbf{77,297} & 33.0\% & 0.75\% & \textbf{0.866} & \textbf{109,045} \\
FinePDFs$^{\dagger}$        & PDF   & 1,230  & 56.5\% & 7.63\%  & 0.522 & 4,682   \\
\textbf{Dolma 3 olmOCR}$^{\dagger}$ & PDF & \textbf{5,056} & \textbf{74.0\%} & \textbf{33.3\%} & \textbf{0.153} & 6,024 \\
\midrule
\multicolumn{7}{l}{\emph{New rows measured for this paper} (E2; seeded
shuffle, seed 20260809; budgets in the caption)} \\
UK Hansard          & transcript & 45,251  & 23.6\% & 0.00\%  & 0.930 & 96,475 \\
YouTube (CC-BY)     & transcript & 1,779   & 0.1\%  & 0.00\%  & 0.996 & 27,550 \\
Pre-1929 books      & books (OCR)& 135,399 & 92.5\% & 10.11\% & 0.413 & 19,147 \\
\bottomrule
\end{tabular}
\caption{Structural markup across public pre-training corpora, tokenizer
\texttt{o200k\_base}. The ten upper rows are the published census, which was
produced by a run that wrote no per-row output file; $\dagger$ marks the six
rows independently re-measured in August 2026 and reported in
Table~\ref{tab:recheck}. Per-row provenance, including the five fields the
original run did not record, is in \texttt{data/census/census-table.json}.
The three lower rows were measured for this paper under a different sampling
rule and are not part of the published ten: each ran at a 25M-token budget
except pre-1929 books, which ran to a 60M-token hard cap and still returned
only 267 documents. Their
intervals are document-level bootstraps with 1,000 resamples, where every
other interval in this paper uses 10,000. \Sw{8{,}192} is computed over
documents of at least 8,192 tokens, because a document shorter than one
window contributes no windows rather than counting as trivially clean
(\S\ref{sec:instrument}); for the web rows that is a small tail of each
corpus, and the median-document and clean-run columns are the ones that
describe the typical document. \Sw{32{,}768} was measured only for the three
new rows, where \S\ref{sec:supply} reports it, and is unmeasured for the
published ten.}
\label{tab:census}
\end{table}

\begin{table}[t]
\centering
\small
\begin{tabular}{lrrrrrl}
\toprule
corpus & docs & any markup & ATX \texttt{\#} & \Sw{8{,}192} & windows & verdict \\
\midrule
C4                  & 17,563 & $4.0 \to 3.8$   & $0.02 \to 0.07$  & $0.889 \to 0.870$ &    23 & agrees \\
FineWeb             & 12,027 & $21.5 \to 20.7$ & $0.12 \to 0.19$  & $0.603 \to 0.613$ &    62 & agrees \\
DCLM                &  6,839 & $27.2 \to 27.6$ & $0.98 \to 0.89$  & $0.505 \to 0.558$ &   138 & agrees \\
Dolma books         &     54 & $33.0 \to 40.7$ & $0.75 \to 0.00$  & $0.866 \to 0.755$ &   983 & best effort \\
FinePDFs            &  1,693 & $56.5 \to 51.6$ & $7.63 \to 2.30$  & $0.522 \to 0.654$ &   610 & best effort \\
Dolma 3 olmOCR      &  1,003 & $74.0 \to 69.3$ & $33.3 \to 20.2$  & $0.153 \to 0.271$ &   657 & \textbf{divergent} \\
\quad at 25M tokens &  2,805 & $74.0 \to 69.3$ & $33.3 \to 21.3$  & $0.153 \to 0.296$ & 2,180 & \textbf{divergent} \\
\bottomrule
\end{tabular}
\caption{August 2026 re-verification of six census rows, printed as
published~$\to$~re-measured. Sampling is head-of-stream at an 8M-token budget
except the last row, which is a 25M-token pass over the same topic-balanced
sample; dataset revisions, shard lists and read orders are recorded per row
in \texttt{data/census/census-table.json}. ``Windows'' is the number of
non-overlapping 8,192-token windows the fresh sample contained, and it is the
reason the three web rows' agreement on $S(8{,}192)$ is weak evidence rather
than strong. The two best-effort rows sample a single shard under a
configuration the original run did not record, so a difference there is not
necessarily drift. The Pile is absent because it cannot be re-verified: the
reachable mirror has Books3, BookCorpus2, OpenSubtitles, YouTube Subtitles
and OpenWebText2 removed, which are precisely the long, low-markup subsets,
making it a different corpus rather than a check on this one.}
\label{tab:recheck}
\end{table}

\begin{figure}[t]
\centering
\includegraphics[width=\linewidth]{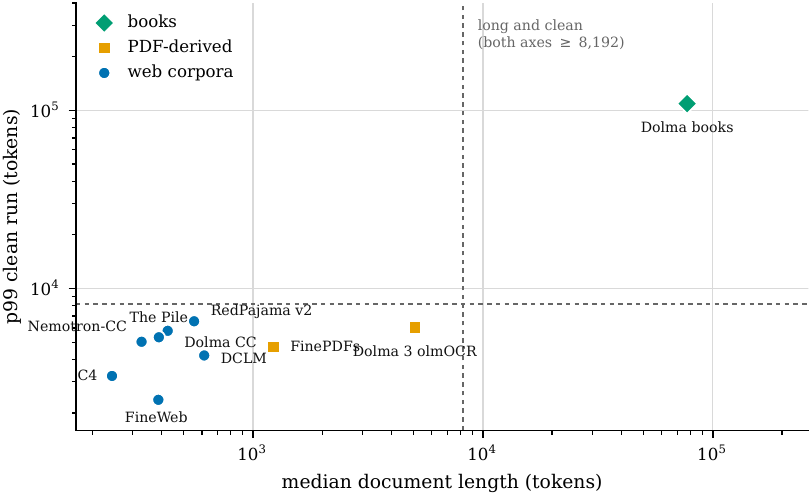}
\caption{The supply problem in one plane. Horizontal: median document length.
Vertical: the 99th-percentile clean run, the longest markup-free spans a
corpus actually delivers. Web corpora are clean and short; PDF-derived
corpora are long and marked; books sit alone in the quadrant where both axes
exceed the training window, and books are the row that left the mixture.}
\label{fig:quadrant}
\end{figure}

\subsection{The supply side}
\label{sec:supply}

Whether the missing format can be resupplied is an empirical question, and
we pre-registered the test before running it. A corpus \emph{supplies} the
missing format at horizon $W$ if and only if its median document is at least
$W$ tokens \emph{and} its $S(W)$ is at least 0.50. Both conjuncts are
required, which is the whole point: the claim under test is that no
available source has both. Three candidates were measured under a seeded
shuffle at a 25M-token budget, except the library-scan books row, which ran
to its 60M-token hard cap and still returned only 267 documents. All three
come from openly licensed collections
\citep{common_pile_2025}, which matters because it turns the supply question
into a shopping list rather than a proposal: that one project has already
assembled both parliamentary records and transcribed audio, and a 242B-token
corpus of digitized library books exists alongside it
\citep{institutional_books_2025}.

UK parliamentary proceedings \textbf{confirm}, at both horizons. Whole
sittings run to a median of 45,251 tokens, survival is 0.930 [0.916, 0.943]
at 8,192 and 0.802 [0.761, 0.839] at 32,768, and the stream carries 1.79
core markup tokens per 10,000. That is a books-class row from a source that
is not books, and it grows.

Creative-Commons YouTube transcripts \textbf{refute}, on length, and this is
the most decision-relevant number in the experiment. Their markup density is
the lowest anywhere in the census, 0.01 core tokens per 10,000 with
$S(8{,}192) = 0.996$, and their median document is 1,779 tokens, an order of
magnitude short of a long-context training window. The fastest-growing transcript
pool in the world is unmarked and unusable for this purpose at the same
time.\footnote{$S(32{,}768)$ for this row is reported as
\texttt{insufficient\_windows}: 56 windows against a pre-registered floor of
200. It must not be used in any decision rule, and it is not.}

Library-scan books \textbf{refute}, on density, and that was the surprise. We
expected them to confirm, because the registration says so: this row was
recorded before measurement as expected to confirm at 32,768 and to be
books-class, and it did neither.
Pre-1929 public-domain books are long, with a median document of 135,399
tokens, and they are not clean: 92.5\% of documents carry detector-visible
markup, the stream runs 33.45 core markup tokens per 10,000, and
$S(8{,}192)$ is 0.413 [0.370, 0.463], falling to 0.142 [0.111, 0.182] at
$S(32{,}768)$. OCR of a printed page reconstructs the
page's furniture. The row is also underpowered against its own
pre-registered floor, at 267 documents rather than 400, and one document, at
7,998,460 tokens, is 13.3\% of the sampled stream on its own. At that length
it is a bound volume set rather than a book, and the row's pooled window
statistics are influenced by it accordingly.

The supply claim holds, and the honest headline is narrower than the one it
replaces: \emph{the long, unmarked speech supply is institutional, not
consumer}. Proceedings qualify; the consumer transcript pool does not. The
verdict rests on one row of three, and that row is the sampling-sensitive
one, so the two facts belong in the same breath: Hansard is flagged
\texttt{sampling\_sensitive} in its own provenance, and reading from the head
of a shard rather than from a seeded shuffle gives
a median document almost eight times lower and $S(8{,}192) = 0.757$ against
0.921, both at the 8M-token sensitivity budget; the row's 25M-token value is
0.930. Corpus statistics of this kind are properties of a sampling rule as
much as of a corpus, which is one more reason the sampling rule belongs on
the card.

\begin{figure}[t]
\centering
\includegraphics[width=\linewidth]{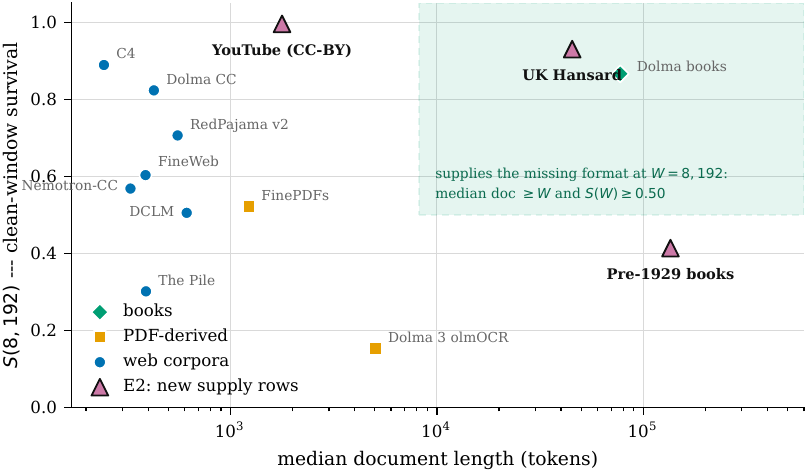}
\caption{The pre-registered supply rule as a region: a corpus qualifies at
horizon $W$ only in the upper-right quadrant, where the median document
reaches $W$ tokens and $S(W)$ reaches 0.50. UK Hansard clears both conjuncts
at 8,192 and at 32,768; Creative-Commons YouTube fails on length despite
having the lowest markup density in the census; pre-1929 library scans fail
on density despite being the longest documents in it.}
\label{fig:supply}
\end{figure}

\subsection{No null level}
\label{sec:nonull}

Across the 50 documents of the ground-truth corpus, the minimum markup
density after serialization is 1.245 core markup tokens per 10,000, in
Conrad; the next three are 2.23, 2.79 and 3.17. \textbf{No document has
zero.} Dispersion between documents passes through conversion essentially
untouched, with a coefficient of variation of 1.190 in the authored ground
truth against 1.162 through Docling and 1.206 through Marker, two of the four
configurations whose dispersion sits within 3\% of the authored value. The
claim holds for four of the five structure-emitting configurations;
\texttt{pymupdf4llm} is 15\% higher at 1.364. So conversion is not homogenizing the spread. What it removes is one
specific value from the support: \emph{absent}. This is why the problem
cannot be reached by mixture weighting, however sophisticated the weighting
is. Reweighting moves mass between levels of a variable, and it cannot
restore a level the variable no longer has.

The volume objection arrives immediately and deserves a direct answer.
Across the fifteen serialized narrative works, structural markup is 664
tokens out of 1,024,826, which is 0.0648\% of the stream and works out to
2.66 structural markers, or 5.31 markup tokens, in an 8,192-token training
sequence. Per work the share ranges from 0.0125\% to 0.2194\%. Nothing in
the mechanism runs on volume: shortcut adoption runs on predictivity and
availability \citep{geirhos_2020,hermann_2024}, and circuit formation runs
on reliability at more than twice its sensitivity to frequency
\citep{aoyama_2026}. A cue present at every boundary and absent everywhere
else would sit at reliability 1.0 and its frequency would be beside the
point; \S\ref{sec:invariance} measures how close the real one comes. The
concession owed in the same breath is that the dose-response curve is
unmeasured. We do not know how much of this material, at what density, is
needed to install or remove a format preference, and our own prior evidence
about how little data installs one cuts against a scarcity story rather than
for it.

\subsection{What conversion actually adds: a pre-registered null}
\label{sec:converter}

We predicted that converters would fabricate structure on prose, and that the
damage would concentrate where organizational variety still lives. We
pre-registered that prediction, and it failed. Over 50 documents with ground
truth derived from machine-readable sources, typeset to real paged media and
run through nine current extraction and conversion configurations, the pooled
imposition loss on prose-like material is $-0.029$ [$-0.047$, $-0.010$] at
$W = 8{,}192$: significantly \emph{negative} on all three prose strata
separately as well as pooled, and negative but not distinguishable from zero
on the structured control, whose interval closes at exactly zero
(Table~\ref{tab:imposition}). Modern converters slightly
under-mark. They are not inventing boundaries; they are faithfully marking
real ones. The registered rule returned \texttt{NULL\_NO\_IMPOSITION}.

That null killed the fabrication mechanism, and what survived it is the
stronger claim. A fabricated cue would at least be noisy, and noise is what
keeps a shortcut from closing. A faithful cue is a trustworthy one, so
accuracy in conversion is not a mitigation of the problem in this paper. It
is the problem.

Six things have to be said about how that verdict was reached. First, the
registered decision rule was close to unsatisfiable at the evaluation point:
$S_\mathrm{marked}(8{,}192)$ for prose-like work is 0.063, and a POSITIVE
verdict required a bootstrap lower bound above 0.10, so the rule would have
needed converters to destroy essentially every clean window that faithful
marking leaves. The amendment log recorded that ceiling before the deciding
run and required the paper to say so in those words. Second, the answer to
that objection was pre-specified in the same entry and is reported here: at
$W = 2{,}048$ the ceiling is 0.547, the rule is satisfiable, and imposition
is still negative pooled, at $-0.011$ [$-0.017$, $-0.005$] for prose against
a control point estimate of $-0.014$ whose interval covers zero. At that
horizon the narrative stratum alone no longer excludes zero, at $-0.006$
[$-0.011$, $+0.001$], and the sentence should not be read as claiming it
does.
Third, the registration promoted Track~B, real publisher and
Internet-Archive PDFs, to primary for the register strata, and the headline
above is Track~A, the self-rendered arm. Track~B cannot support the
\emph{imposition} estimand for prose: its eleven real scanned narrative books
have no ground-truth marked stream to compare against, so the prose contrast
is undefined there. It can support the emission ratio, which needs only the
authored heading count, and that number is reported below rather than
omitted. The deviation itself was not recorded in the amendment log at the
time and is disclosed
here. Fourth, the study contains no vision-language converter: olmOCR and
MinerU were provisioned and produced no rows, so a null measured on nine
non-visual configurations is being carried into an argument whose own
genealogy makes visual conversion the frontier. Fifth, and in the other
direction, the pre-registration is better than the usual: version one was
frozen before any data existed, and the one amendment that changed the
primary outcome documents the exact data state at the moment it was made,
eleven scored cells over two documents with zero technical documents
converted, which makes an outcome-driven amendment impossible for the
quantity the new rule turns on.\footnote{Ground truth for the 50 documents
comes from Gutenberg plain text, arXiv \LaTeX{}, RFC XML, and Markdown and
RST sources. Registers are R0 narrative (15), R1 sustained argument (6), R2
mixed (6) and R3 structured technical writing (23), the last serving as the
control. Nine tool configurations were run; the imposition estimate pools
over the five that emit structure, the other four being flatteners for which
survival is 1.0 by construction. The pooled negative is carried by three of
those five: Marker's imposition on prose is exactly $0.000$ and Docling's
interval, $-0.032$ [$-0.065$, $+0.003$], covers zero.}

Sixth, the registered rule admits four verdicts and this study returns two of
them. The second is
\texttt{NULL\_NO\_CONCENTRATION}, and it fires in the direction opposite
to the one we predicted: the prose-minus-control contrast is $-0.025$
[$-0.045$, $-0.005$], so the loss is not merely unconcentrated on prose but
significantly \emph{smaller} there than on the structured control, where we
had predicted it would be larger. The registered monotonicity secondary is
likewise not satisfied, at $-0.005$ [$-0.013$, $+0.003$], with the stratum
ordering non-monotone at both horizons. And the ceiling-free relative
estimand the amendment log pre-specified at every $W$ is computed and
reported here for the first time: $-0.46$ [$-1.08$, $-0.14$] at
$W = 8{,}192$ and $-0.020$ [$-0.033$, $-0.009$] at $W = 2{,}048$. All three
run with the null rather than against it, which is exactly why they are easy
to leave out and worth printing.

One quantity from the same runs sharpens the null rather than softening it,
and it has to be reported on three surfaces because it moves between them.
Counted as core-markup spans emitted per authored heading and pooled over all
fifty documents, the five structure-emitting configurations run between 1.19
and 1.97 on PDFs rendered from the marked source, against 1.84 spans per
authored boundary in the faithful marked serialization itself, or 1.85
against the 2,599 authored headings the converter join
counts.\footnote{Five configurations are three tools: \texttt{docling} and
\texttt{docling\_furniture} differ only in content layer and emit identical
span counts, and the two \texttt{pymupdf4llm} configurations bracket the
range.} The converters are tracking the source rather than inventing on top
of it. On narrative alone the pooled figure is misleading, because it is
carried by the technical stratum's list markers: there the same tools run
between 1.01 and 1.08, and rendering the \emph{flat} source drops them to
between 0.09 and 0.95. The navigation overlay is therefore partly, not
wholly, manufactured from the page. On structured material it is reproduced
at essentially the same rate with or without source markup; on prose,
removing the markup cuts it on every converter, by a quarter or more on all
but the layout mode and by a tenth there.

Track~B runs higher still on the register it covers, and we report it because
the registration made that track primary. Its eighteen real publisher and
Internet-Archive PDFs are all technical writing, and over them the same
configurations run between 1.27 and 4.43 spans per authored heading, four of
the five at 2.52 or above, against 1.56 to 2.63 on the technical documents we
typeset ourselves. Docling emits 74\% more on real paged media than on our
own rendering of the same register and the layout-aware
\texttt{pymupdf4llm} 83\% more, while Marker emits 4\% less: the
over-emission is real, and it is tool-specific rather than universal. Three
limits travel with that comparison and none of them removes it. The two arms
are different document sets at the same register rather than the same
documents rendered twice, so it compares populations and not pairs.
Track~B's scanned-book arm is nineteen documents, sixteen of them prose, and
the single tool run over them emits no structural spans at all, so the
register this paper's proposal is about is not covered there by any tool that
emits structure. And the rendering path of the self-rendered
PDFs was not independently inspected here, so we report the ratios and not a
mechanism. The direction is what matters for the null above: under-marking is
a property of documents we typeset ourselves, and on real paged media two of
these three tools over-mark substantially.

\begin{table}[t]
\centering
\small
\begin{tabular}{lrrl}
\toprule
stratum & $n$ & imposition loss at $W=8{,}192$ & 95\% CI \\
\midrule
R0 narrative               & 15 & $-0.015$ & \ci{-0.034}{-0.001} \\
R1 sustained argument      &  6 & $-0.054$ & \ci{-0.095}{-0.010} \\
R2 mixed                   &  6 & $-0.037$ & \ci{-0.073}{-0.008} \\
R3 structured (control)    & 23 & $-0.004$ & \ci{-0.011}{\phantom{-}0.000} \\
\midrule
\textbf{prose-like (R0 + R1), pooled} & 21 & $\mathbf{-0.029}$ & \ci{-0.047}{-0.010} \\
\multicolumn{4}{l}{\quad registered verdict: \texttt{NULL\_NO\_IMPOSITION}} \\
\midrule
\multicolumn{4}{l}{\emph{$W=2{,}048$ companion, where the registered rule is satisfiable}} \\
prose-like, pooled         & 21 & $-0.011$ & \ci{-0.017}{-0.005} \\
control                    & 23 & $-0.014$ & \ci{-0.036}{\phantom{-}0.012} \\
\bottomrule
\end{tabular}
\caption{Imposition loss from nine converter configurations over 50
ground-truth documents, cluster bootstrap (seed 20260805, $n=10{,}000$).
Imposition loss is $S_\mathrm{marked} - S_\mathrm{converted}$, so a negative
value means the converter emitted \emph{less} structure than the authored
source and left more clean windows standing. At $W = 8{,}192$ the estimand's
own ceiling is $S_\mathrm{marked} = 0.063$ against a registered POSITIVE
threshold of 0.10, which made the rule close to unsatisfiable at that
horizon; the $W = 2{,}048$ companion, where the ceiling is 0.547 and the rule
is satisfiable, is the pre-specified answer to that objection and returns the
same sign.}
\label{tab:imposition}
\end{table}

\subsection{Notation invariance and cue reliability}
\label{sec:invariance}

The fourth measurement is the one that makes the obvious fix wrong. Strip the
\texttt{\#} but leave the chapter title as a short standalone line between
blank lines, and the notation-blind detector of \S\ref{sec:instrument} finds
the boundaries in essentially the same places (Table~\ref{tab:invariance}).
On narrative the two readings differ by $-0.001$, $-0.010$, $-0.004$ and
$0.000$ at 1,024 through 8,192 tokens; on sustained argument they differ by
at most $+0.007$. We print all four registers rather than the two the
argument needs, because the exceptions are informative: mixed material
differs by $+0.031$ at 8,192, and structured technical writing by $-0.015$
and $-0.018$ at the two shortest horizons, where the detector reads
enumerated skeletons the syntax check misses.

Flattening the notation, on its own, changes almost nothing about where
boundaries are detectable, because the typographic form is itself a perfectly
good cue: a short bare line with whitespace around it. This zero is a result
and not a nuisance, and it dictates the shape of any intervention. What has
to be attacked is not the sigil's presence but its \emph{reliability}.
Removing one notation while leaving another equally reliable one in place
merely swaps one shortcut for another, which is the finding that sends
\S\ref{sec:pureframe} to the level of the announcement rather than the level
of the marker.

The hinge of the whole mechanism is that the cue is reliable, so we
pre-registered a test of that sentence rather than continuing to assert it.
E5 matches every structural sigil in a serialization against the authored
boundaries of the same document, one to one, and reports precision and recall
per stratum (Table~\ref{tab:reliability}). The pre-registered verdict is
\texttt{UNDERMINED} as stated. The rule required precision at or above 0.99
with a lower bound at or above 0.95 in narrative and in sustained argument,
and sustained argument returns 0.765 [0.521, 1.000] against notation and
0.722 [0.485, 0.977] against typography. Narrative, which is where the
missing format lives, holds: 0.994 [0.984, 1.000] and 0.956 [0.933, 0.977].

Two things about that failure are worth more than the verdict. It is carried
by a single document: Arnold's \emph{Culture and Anarchy}, whose Project
Gutenberg text carries a numbered endnote apparatus that the syntax detector
reads as ordered-list markers, contributing 31 of the false positives on its
own. Dropping it restores precision to 1.000 and 0.951, which we report as
post hoc and not as an amendment to the rule; it is the same false-positive
mode that accounts for the residue in the flat pack of
\S\ref{sec:pureframe}. And the pooled row is dominated by structured
technical writing at 0.418, where a sigil is usually marking a list rather
than announcing a discourse boundary, which is not a failure of the cue so
much as a reminder of what the cue is for.\footnote{One arm's recall is
definitional and should not be read as a finding: the marked serializer
writes a sigil at every heading, so the syntax detector cannot miss one and
its recall is 1.000 by construction. Precision on that arm is not
definitional, because body text can mimic markup, and it is the informative
half. The structural arm on flat text is the one whose recall carries
information, and it comes in at 0.967 in narrative.}

Two secondary verdicts from the same run are worth a sentence, because they
turn the converter null of \S\ref{sec:converter} into a stronger statement.
Recall returns \texttt{PERFECT\_RECALL} on the notation arm and
\texttt{CONSISTENT\_WITH\_UNDER\_MARKING} on the typographic one, which
converts ``converters do not fabricate'' into ``converters do not fabricate
\emph{and} they under-mark''. The detector-agreement check passes, with the
recall gap between the two arms at 0.033 in narrative and 0.050 in sustained
argument against a threshold of 0.05, so the second of those sits exactly on
the line rather than under it.

The sentence we retire is that the cue never lies. What replaces it is scoped
and more useful. In narrative, the material this paper's proposal is actually
about, a structural sigil announces an authored boundary at precision 0.994
[0.984, 1.000] against the manifests as built, and at 0.967 once the
thirty-two heading blocks our own strata file documents as typographic
apparatus are removed from the ground truth.\footnote{A ground-truth defect
worth recording alongside it: the strata file's NOTE 3 states fifteen such
blocks across five documents, while its own enumeration resolves to
thirty-two across eight. We use the enumeration, because it is the auditable
list.} We report both because the registration required both, and the
adjusted figure is the one that decides the band: 0.967 falls in the
registered SUPPORTED, WEAKENED range rather than the SUPPORTED AS STATED one,
whose pre-stated consequence is that this paper writes \emph{near-perfect}
reliability rather than perfect reliability everywhere the claim appears,
including the abstract. Near-perfect is still what a shortcut runs on.
Elsewhere the sigil is doing other work and its reliability is a measured
quantity between 0.42 and 0.84 on either reading. A
mechanism that only holds where the paper applies it is a narrower claim than
the one we started with, and it is the one the data supports.

\begin{table}[t]
\centering
\small
\begin{tabular}{lrrrr}
\toprule
register & $W$ & $S$, Markdown syntax & $S$, structural detector on flat text & difference \\
\midrule
narrative           & 1,024 & 0.701 & 0.699 & $-0.001$ \\
narrative           & 2,048 & 0.472 & 0.462 & $-0.010$ \\
narrative           & 4,096 & 0.202 & 0.198 & $-0.004$ \\
narrative           & 8,192 & 0.026 & 0.026 & $+0.000$ \\
\addlinespace
sustained argument  & 1,024 & 0.846 & 0.847 & $+0.002$ \\
sustained argument  & 2,048 & 0.703 & 0.711 & $+0.007$ \\
sustained argument  & 4,096 & 0.496 & 0.500 & $+0.004$ \\
sustained argument  & 8,192 & 0.140 & 0.143 & $+0.003$ \\
\addlinespace
mixed               & 1,024 & 0.484 & 0.465 & $-0.019$ \\
mixed               & 2,048 & 0.275 & 0.272 & $-0.003$ \\
mixed               & 4,096 & 0.113 & 0.113 & $+0.000$ \\
mixed               & 8,192 & 0.010 & 0.042 & $+0.031$ \\
\addlinespace
structured          & 1,024 & 0.040 & 0.025 & $-0.015$ \\
structured          & 2,048 & 0.018 & 0.000 & $-0.018$ \\
structured          & 4,096 & 0.010 & 0.000 & $-0.010$ \\
structured          & 8,192 & 0.000 & 0.000 & $+0.000$ \\
\bottomrule
\end{tabular}
\caption{Clean-window survival measured two ways: against Markdown syntax in
the marked serialization, and against a notation-blind structural detector
run over the flattened text. On prose the two agree to within a hundredth at
every horizon, the largest gap being 0.010 on narrative at 2,048 tokens, so
the boundary is readable from typography alone and flattening the notation
buys nothing. All four registers are printed, including the two where the
readings come apart.}
\label{tab:invariance}
\end{table}

\begin{table}[t]
\centering
\small
\setlength{\tabcolsep}{2.5pt}
\begin{tabular}{llrrlrl r}
\toprule
stratum & detector & bounds & sigils & precision & adj. & recall & $F_1$ \\
\midrule
R0 narrative          & notation   &   334 &   336 & 0.994 \ci{0.984}{1.000} & \textbf{0.967} & 1.000 \ci{1.000}{1.000} & 0.997 \\
R0 narrative          & typography &   334 &   338 & 0.956 \ci{0.933}{0.977} & \textbf{0.935} & 0.967 \ci{0.936}{0.993} & 0.961 \\
\addlinespace
R1 sustained argument$^{*}$ & notation   &   101 &   132 & 0.765 \ci{0.521}{1.000} & \textbf{0.606} & 1.000 \ci{1.000}{1.000} & 0.867 \\
R1 sustained argument$^{*}$ & typography &   101 &   133 & 0.722 \ci{0.485}{0.977} & \textbf{0.594} & 0.950 \ci{0.871}{1.000} & 0.821 \\
\addlinespace
R2 mixed$^{*}$        & notation   &   715 &   849 & 0.842 \ci{0.801}{0.899} & 0.840 & 1.000 \ci{1.000}{1.000} & 0.914 \\
R2 mixed$^{*}$        & typography &   715 & 1,009 & 0.619 \ci{0.521}{0.805} & 0.617 & 0.874 \ci{0.729}{0.979} & 0.725 \\
\addlinespace
R3 structured         & notation   & 1,458 & 3,484 & 0.418 \ci{0.337}{0.526} & 0.418 & 1.000 \ci{1.000}{1.000} & 0.590 \\
R3 structured         & typography & 1,458 & 2,528 & 0.562 \ci{0.430}{0.711} & 0.562 & 0.975 \ci{0.958}{0.987} & 0.713 \\
\midrule
all 50 documents      & notation   & 2,608 & 4,801 & 0.543 \ci{0.448}{0.660} & 0.537 & 1.000 \ci{1.000}{1.000} & 0.704 \\
all 50 documents      & typography & 2,608 & 4,008 & 0.615 \ci{0.514}{0.726} & 0.609 & 0.945 \ci{0.904}{0.977} & 0.745 \\
\bottomrule
\end{tabular}
\caption{The mechanism's hinge as a measurement: how reliably a structural
sigil announces an authored boundary, matched one to one against ground
truth on the 50-document corpus, cluster bootstrap by document (10,000
resamples, seed 7). Boundaries are headings and section breaks. The
pre-registered verdict on ``the cue never lies'' is \texttt{UNDERMINED} as
stated, because sustained argument falls below the 0.99 threshold; narrative,
which is the material the paper's proposal addresses, satisfies it as-is and
lands in the SUPPORTED, WEAKENED band once adjusted. Precision is reported
twice as the registration required: as-is, and in the \emph{adj.} column with
the thirty-two heading blocks the strata file documents as typographic
apparatus removed from the boundary set. The
notation arm's recall is 1.000 by construction and carries no information;
its precision does. $^{*}$ fewer than ten documents: point estimates stand,
intervals are underpowered, and strata are not merged to manufacture power.}
\label{tab:reliability}
\end{table}

\section{The one stage where the sequence is the document}
\label{sec:slot}

Where in training would any of this bind? The answer has a shape that is
easy to miss and hard to unsee afterwards.

\textbf{Over 95\% of pre-training tokens are consumed at sequence lengths of
8,192 or less.} Llama~3 ran 15.6T of its roughly 16.4T tokens at 8,192, with
a long-context extension of about 800B \citep{llama3_2024}. DeepSeek-V3
pre-trained 14.8T tokens at 4,096 and extended context with about 126B tokens
of YaRN, under one percent of the budget
\citep{deepseekv3_2024}.\footnote{The two YaRN phases are exactly equal by
construction: 1,000 steps at $32{,}768 \times 1{,}920$ and 1,000 at
$131{,}072 \times 480$, and $1{,}920 \times 32{,}768 = 480 \times 131{,}072$.}
Qwen3 ran roughly 35T of its 36T at 4,096 \citep{qwen3_2025}. Olmo~3's 7B ran
5.93T of pre-training and 100B of midtraining at 8,192, then 50B at 65,536
\citep{olmo3_2026}. The worst case among those four families is 95.12\% and
the token-weighted pool is 97.31\%, so ``over 95\%'' is not a rounded claim.
It is the floor.

At the main stage a training sequence is not a document. It is a packed
window: several short documents concatenated, or an arbitrary slice from the
middle of a long one, frequently straddling a document boundary. At that
stage the attention mask usually does not even separate the neighbours.
Llama~3 reports that document-separating masks had ``limited impact'' in
standard pre-training while being ``important in continued pre-training on
very long sequences'' \citep{llama3_2024}, and DeepSeek-V3 packs documents
but ``do[es] not incorporate cross-sample attention masking''
\citep{deepseekv3_2024}.

There is a statistical-mechanics way to say what such a stage can and cannot
teach. Mutual information between tokens of natural text decays as a power
law, and that decay is caused by hierarchical document structure, with
correlations flowing from the high organizational levels down into the words
\citep{alvarez_lacalle_2006,altmann_2012}; transformers are preferred in part
because they capture slow power-law dependence where recurrent models cannot
\citep{shen_2019_mi,lin_tegmark_2017}.\footnote{Two glosses worth keeping
exact: \citet{sarkar_howard_2019} states the power-law decay as background to
a different contribution, and \citet{mikhaylovskiy_2023} measures
autocorrelation rather than mutual information, concluding that
Markov-behaving autoregressive models ``may have limitations when applied to
long texts''.} Set against that: authored boundaries arrive every 3,078
tokens of narrative on average and every 2,392 tokens at the median, which is
worse; sequences are capped at 8,192; and every web corpus's clean runs are
exhausted by 6,500. Planning-horizon work points the same way, finding that
models pre-cache features for future tokens where the data has genuine future
structure and that pre-caching rises with scale \citep{wu_2024_precaching}.
Packing sharpens the picture: sequences cut from the marked serialization of
our narrative corpus survive at 0.033 at $L = 8{,}192$, against a ceiling of
0.888 for the same books flattened.\footnote{The 0.888 residual is entirely
document-straddling rather than markup: markup-only survival for the flat arm
is 1.000, and 11.2\% of 8k sequences cross a book boundary. Full simulation,
60 rows, in \texttt{data/analysis/packing.json}.}

The long-context stage inverts every one of those properties. Sequence length
exceeds the median book, and intra-document masking is switched on. Olmo~3's
released training scripts run the main stages at
\texttt{DEFAULT\_SEQUENCE\_LENGTH = 8192} with no masking flag, while both
the 7B and the 32B long-context scripts read \citep{olmo_core_2026}:

\begin{lstlisting}
DEFAULT_SEQUENCE_LENGTH = 65536
...
generate_doc_lengths=True,  # enables intra-document masking
\end{lstlisting}

At this stage, and only at this stage, \textbf{the training sequence is the
document}: one work, entire, in reading order, with attention confined to it.
It is the only point in the pipeline where a gradient can flow through the
experience of tracking one document's structure across tens of thousands of
tokens.

What that stage is fed is therefore the whole question, and the public record
answers it precisely. Olmo~3's long-context data pool is 639B tokens and is,
to a rounding error, entirely olmOCR-converted science PDFs: every length
bucket from 8K to 1M and beyond is a PDF row, and there is no books row, no
transcript row, and no long-form prose row of any kind. The 50B stage
actually trained draws 33.9\% from that pool and replays midtraining data for
the other 66.1\% \citep{olmo3_2026}. The consequence is sharper than the
headline number. Every long document the model meets at 65,536 tokens with
intra-document masking on comes from the corpus with the lowest clean-window
survival in our census, published at $S(8{,}192) = 0.153$ and re-measured at
0.271, which is the lowest figure either way. The remaining two-thirds is the
ordinary short-document mix, packed, which is the stage the long-context
stage was supposed to be different from.

Nothing in this is anyone's error, and the field has noticed the scarcity
without naming its cause. SmolLM3's team reports that upsampling long-context
data such as code repositories, books and long web pages ``didn't further
boost performance on RULER and HELMET benchmarks'' \citep{smollm3_2025}, and
a 2026 mid-training line builds long-context supervision out of
cross-repository code because, in its authors' words, the existing
long-document sources are ``finite resources and often scarce in
long-distance dependencies'' \citep{octolong_2026}. Meanwhile every
long-context data method in current use, whether selection, synthesis or
supervision reweighting, keys on long-range information gain, attention
statistics or predictive uncertainty
\citep{longrange_info_2026,longattn_2025,ladm_2025,entropylong_2025,
nextlong_2025,effective_context_2026}, and none of those criteria conditions
on how the candidate is notated. The variable is invisible to the
tools that choose what fills the slot, which is also why the intervention is
cheap: it lives in a stage that is under one percent of the budget, in
training runs whose infrastructure is already configured for it.

\begin{figure}[t]
\centering
\includegraphics[width=\linewidth]{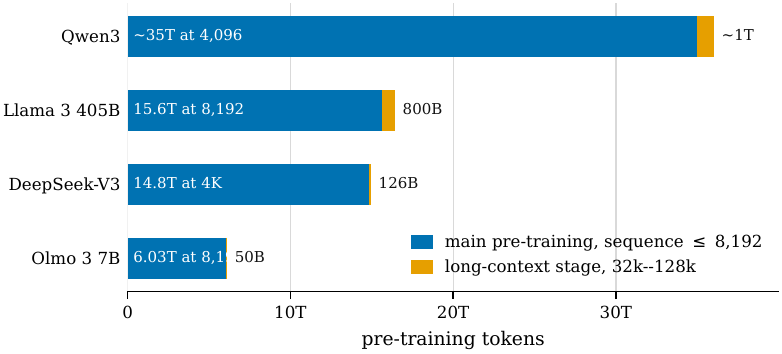}
\caption{Where the tokens go. For each family, the main pre-training stage
(sequence length at most 8,192, no intra-document masking) against the
long-context stage (32K to 128K, masking on). The stage with the right
mechanics for long-range structure, where the sequence is the document, is a
sliver of the budget, which is precisely what makes it cheap to fill
deliberately.}
\label{fig:tokenslot}
\end{figure}

\section{Front II: what readers use}
\label{sec:front-readers}

Front~I measured what the corpora carry. This front measures what a trained
reader does with it, on text where the notation is the only thing that
moves. Twenty-one narrative and sustained-argument works are serialized from one
block structure into three arms. The \emph{marked} arm writes the genuine
structure as Markdown; the \emph{flat} arm keeps every word, drops the
notation, and leaves the heading text standing as a bare line; the
\emph{silent} arm deletes the announcement line and puts nothing in its
place. Marked against flat isolates the sigil, flat against silent isolates
the announcement. Targets are ordinary interior prose, never headings,
stratified by distance from the last authored boundary and byte-identical
across arms: 169 across the 21 works, and 88 in the silent arm, which exists
only for the twelve works whose chapter heads are contentless labels. Each
target is scored with 768 tokens of preceding context and again with 7,936,
and information gain is the difference, in millinats per byte. Intervals are
cluster bootstraps by work, 10,000 resamples, seed 7.

\begin{figure}[t]
\centering
\includegraphics[width=\linewidth]{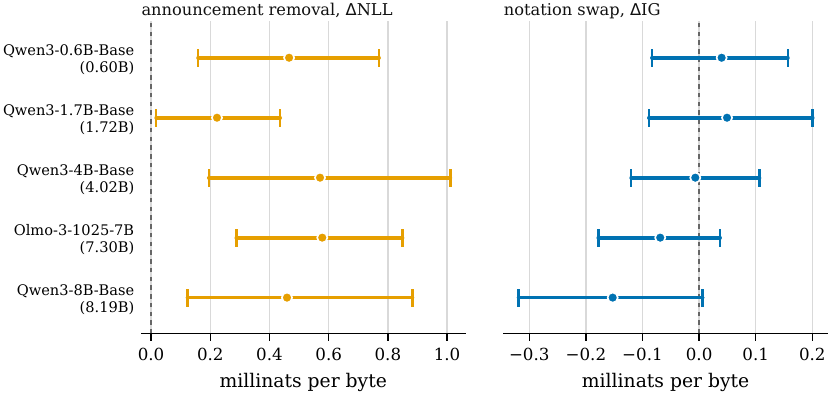}
\caption{The two contrasts across the reader ladder, with 95\% cluster
bootstrap intervals. Left: deleting the announcement raises prediction
difficulty at every completed scale, including the cross-family reader.
Right: swapping the notation on the same boundary moves nothing, at every
completed scale, inside a bound under two percent of measured information
gain. All five readers are complete.}
\label{fig:probescales}
\end{figure}

The published measurement was one 0.6-billion-parameter reader, which was
this work's own first-listed limitation. It is now five, spanning 0.60B to
8.19B and two independent pre-training pipelines
(Table~\ref{tab:crossscale}, Fig.~\ref{fig:probescales}). The anchor-set
hash is identical for all five and equals the published run's, so the
comparison is exactly paired by construction rather than by post-hoc
alignment, and recomputing the 0.6B row through the cross-scale code
reproduces every published estimate to three decimals.

\begin{figure}[t]
\centering
\includegraphics[width=\linewidth]{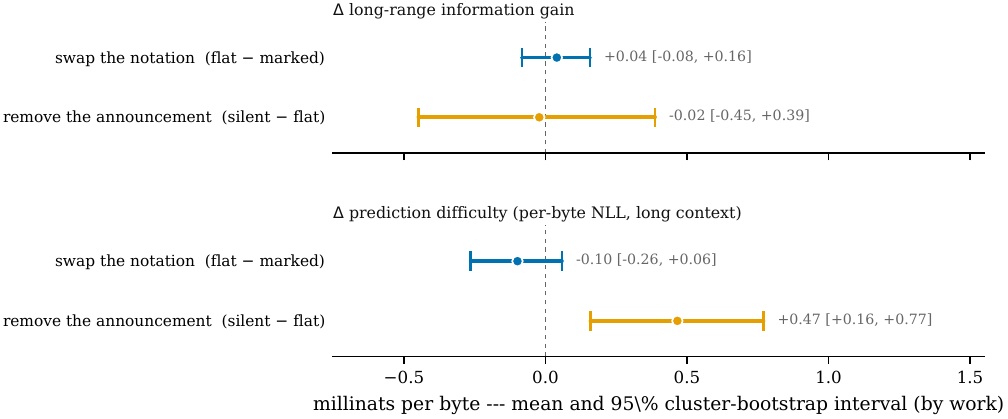}
\caption{The three-arm probe at the published 0.6B reader. Swapping the
notation moves nothing; removing the announcement makes the following prose
measurably harder to predict, and none of that loss is recovered from
long-range context. Millinats per byte, cluster bootstrap by work.}
\label{fig:probe}
\end{figure}

\textbf{R1, the deficit persists across scales: satisfied at five of five.}
Deleting the announcement makes the prose that follows it harder to predict
at every scale measured. The rise is $+0.466$ [$+0.159$, $+0.770$] millinats
per byte at 0.6B, $+0.223$ [$+0.017$, $+0.436$] at 1.7B, $+0.571$
[$+0.195$, $+1.011$] at 4B, $+0.578$ [$+0.288$, $+0.849$] at the
cross-family reader, and $+0.459$ [$+0.123$, $+0.883$] at 8B, the largest
reader of the five. Each lower bound
stays above zero at three bootstrap seeds. The finding is no longer a single
measurement: it holds over a $13.7\times$ span of scale and across two
pre-training pipelines that share no data decisions. It also has a published
precedent, which we extend rather than compete with. ChapterBreak handed
long-range language models a narrative segment ending at a chapter boundary
and asked them to pick the true next chapter out of same-narrative
negatives, and reported that they ``fail to effectively leverage long-range
context, substantially underperforming a segment-level model trained
directly for this task'' \citep{chapterbreak_2022}. That is this front's
question four years earlier: their task is discriminative and supervised
over candidate continuations, ours is a paired likelihood contrast on
identical bytes with the announcement present or deleted, and the curve
below adds a spatial profile their design cannot see.

\textbf{R3, the notation swap is a zero everywhere: satisfied, and tight.}
Every reader's notation contrast contains zero and every one is tight
against the pre-registered bound, the widest interval endpoint being 0.65\%,
0.91\%, 0.55\%, 0.89\% and 1.68\% of that reader's measured information gain
against a threshold of 2\%. The largest reader is the loosest of the five and
is still inside the bound. This is the most misreadable result in the paper, so we
interpret it here as well as in the abstract. A zero does not say that
notation is unimportant. It says the sigil is not the operative cue: present
the chapter head as \texttt{\#\# CHAPTER IV.} or as \texttt{CHAPTER IV.} and
a trained reader does the same thing with the prose that follows, at every
scale we ran. What moves the reader is whether the line is there at all,
which is why \S\ref{sec:pureframe} specifies its intervention over
announcement presence rather than over notation. One corollary matters for
curation: information-gain scores of the kind used to select long-context
training data are insensitive to serialization across the scales tested.
The swap's point estimate declines with scale inside the null, from
$+0.05$ at 1.7B to $-0.15$ at 8B, and at 8B the interval's upper bound is
$+0.006$: one more doubling in the same direction and the registration's
named Reversal branch, the marked arm using \emph{more} long-range
information, would fire. The five readers share one anchor set, so this is
one correlated trend rather than five findings; the third-pipeline reader
of the replication extension below sits on the positive side of the null,
and the training-time test is where it resolves.

One sign inside that null is worth stating, because a
reader will notice it before we point at it. The other half of the notation
contrast, the raw difference in prediction difficulty rather than in
long-range gain, has a negative point estimate at four of the five readers,
running
from $-0.04$ to $-0.10$ millinats per byte: flat text is marginally easier
than marked text, not harder. The 8.19B reader is the exception, at $+0.028$
[$-0.094$, $+0.163$]. Every interval covers zero, and the closest to
excluding it is the cross-family reader's $[-0.206, +0.014]$. Because the
five readers share one anchor set, the five signs are not independent draws,
so this is one correlated near-null rather than five of them. We report the
signs anyway, because four of the five run the same way and because that
direction points away from a suppression story rather than toward one.

\textbf{R2, larger readers reconstruct: fired, and it cuts against this
paper.} The rule asked whether a bigger reader answers a deleted
announcement by reaching further back into context, and its first branch
fired: at 1.72B the announcement information-gain contrast is $+0.270$
[$+0.010$, $+0.521$], an interval excluding zero from above. Under the
pre-registered rules that is a headline finding in its own right, and what
it says is that the inference may be present in larger readers and missing
only in small ones. We report the rule as fired and the robustness picture
together, because the second is what a reader needs in order to use the
first. The effect replicates on none of the three larger readers: 4B returns
$+0.028$
[$-0.327$, $+0.404$], the 7.30B cross-family reader $+0.612$ [$-0.100$,
$+1.418$] and 8B $+0.151$ [$-0.109$, $+0.394$], all covering zero. Four
readers were eligible under the rule and
one fired, and the same-family sequence is not ordered in scale: $-0.022$,
$+0.270$, $+0.028$ and $+0.151$ from 0.6B to 8B. The deciding bound is
$+0.010$; across three bootstrap seeds it falls as low as $+0.003$, so the
rule fires at every seed we ran, with a margin above zero at its closest
seed smaller than the $\pm 0.005$ reproducibility envelope this project
fixed for the probe before the run. The attenuation branch did not fire at all, because the deficit is not
monotone on the same-family ladder. We report R2 as satisfied because the
rule was written in advance and it fired, and we report in the same place
that a single marginal interval out of four eligible tests is the weakest
form in which it could have fired. The honest sentence is that
reconstruction is not ruled out at reader scales above 1.7B rather than that
it has been shown, and that the matched
training run of \S\ref{sec:program} is what settles it.

\textbf{R4, the determinism gate: passes per reader.} Anchors whose visible
context contains no boundary in either arm receive byte-identical inputs, so
their contrast must be exactly zero, and all eleven return exactly zero,
bitwise, on every completed reader. One descriptive quantity travels
alongside, labelled descriptive because it was not pre-registered: mean
information gain on the marked arm falls as the reader grows, 23.98 to 22.05
to 21.87 to 18.99 millinats per byte up the Qwen3 ladder, with 19.94 at the
cross-family reader, which sits above the largest Qwen3 reader rather than
below it, so the ordering across families is confounded and only the
within-family decline is read here. Larger readers gain less from the extra
7,168 tokens
because they already predict the short-context continuation better, which is
why the announcement contrast is reported in absolute units.

\begin{table}[t]
\centering
\footnotesize
\setlength{\tabcolsep}{4pt}
\begin{tabular}{lrllrl}
\toprule
reader & params & announcement $\Delta$NLL & notation $\Delta$IG & mean IG & nulls \\
\midrule
Qwen3-0.6B-Base & 0.60B & $+0.466$ \ci{+0.159}{+0.770} & $+0.040$ \ci{-0.083}{+0.157} & 23.98 & 11/11 \\
Qwen3-1.7B-Base & 1.72B & $+0.223$ \ci{+0.017}{+0.436} & $+0.049$ \ci{-0.089}{+0.201} & 22.05 & 11/11 \\
Qwen3-4B-Base   & 4.02B & $+0.571$ \ci{+0.195}{+1.011} & $-0.007$ \ci{-0.120}{+0.107} & 21.87 & 11/11 \\
Olmo-3-1025-7B  & 7.30B & $+0.578$ \ci{+0.288}{+0.849} & $-0.069$ \ci{-0.178}{+0.037} & 19.94 & 11/11 \\
Qwen3-8B-Base   & 8.19B & $+0.459$ \ci{+0.123}{+0.883} & $-0.153$ \ci{-0.319}{+0.006} & 18.99 & 11/11 \\
\midrule
\multicolumn{6}{l}{\emph{pre-stated verdicts} (\texttt{EXPERIMENTS.md} \S1.6, evaluated mechanically)} \\
R1 deficit persists   & \multicolumn{5}{l}{\textbf{SATISFIED} at 5 of 5: lower bound above zero for every reader} \\
R2 larger readers reconstruct & \multicolumn{5}{l}{\textbf{SATISFIED via (a)} at 1.7B only: $\Delta$IG $+0.270$ \ci{+0.010}{+0.521}} \\
R3 notation null replicates & \multicolumn{5}{l}{\textbf{SATISFIED} at 5 of 5 and tight: $0.55$--$1.68\%$ of mean IG (bound $2\%$)} \\
R4 determinism gate   & \multicolumn{5}{l}{\textbf{PASS} at 5 of 5: 11/11 bitwise zero per reader} \\
\bottomrule
\end{tabular}
\caption{The probe across the reader ladder, in millinats per byte, cluster
bootstrap by work (10,000 resamples, seed 7). Announcement $\Delta$NLL is
the rise in prediction difficulty when the announcement line is deleted;
notation $\Delta$IG is the change in long-range information gain when the
same boundary is written as a bare line instead of a Markdown heading. The
anchor set is identical across all five readers and identical to the
published run, 169 targets in 21 works with the same
\texttt{anchor\_set\_sha256}, so the comparison is exactly paired by
construction and no contrast moves by a bootstrap draw under the matched-set
rule. The ladder is five \emph{completed} readers spanning $13.7\times$ of
scale, 0.60B to 8.19B, and two pre-training pipelines; the fifth,
Qwen3-8B-Base, completed after the writer-side runs and enters on the same
anchor set as the other four. Weights are bf16 with float32 scoring
throughout. Verdicts are the pre-registered rules applied mechanically over
the complete ladder.}
\label{tab:crossscale}
\end{table}

The silent arm is not an ablation of a formatting nicety. It is the format
this paper ships (\S\ref{sec:pureframe}), applied to the twelve pure-label
works: every structural announcement deleted, nothing put in its place. The
announcement column of Table~\ref{tab:crossscale} is therefore the
per-reader cost, in millinats per byte, of reading the pure frame, and R1
says that cost is real at every scale we measured. It is the inference
demand the format exists to reinstate. The probe deletes pure-label heads
only, because deleting a titled head would confound announcement removal
with content removal, so these numbers bound the format's effect on its
cleanest subset rather than on the full specification.

An anchor-level average says the announcement costs something. It does not
say where the cost lives, and a local re-orientation charge and an
accumulating debt are different claims about what the pure frame does to a
training experience. The last experiment runs the same three arms over one
entire book: every token of all three serializations of \emph{Moby-Dick},
scored with at least 4,096 tokens of contiguous same-arm context, 860,715
scored tokens over 209 windows, paired across arms by paragraph and by
position within the paragraph. The pairing is exact rather than approximate,
and the pure-against-flat comparison pairs 100.000\% of its tokens.

\begin{figure}[t]
\centering
\includegraphics[width=\linewidth]{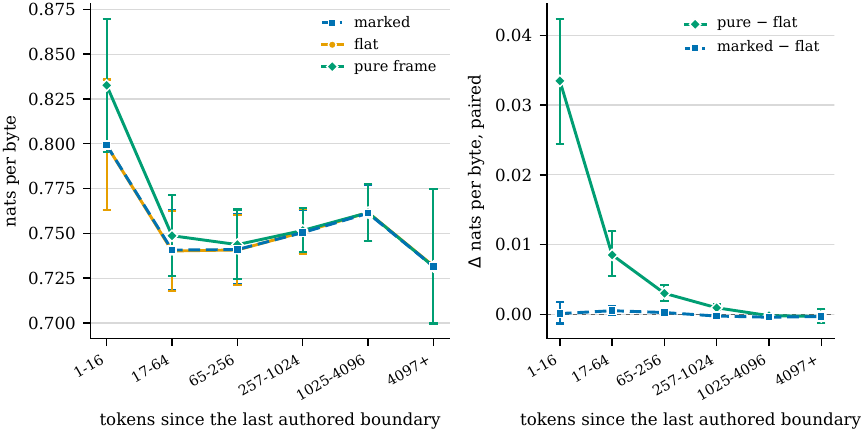}
\caption{The announcement deficit in space, over one entire book, at
Qwen3-1.7B-Base. Left: nats per byte against distance in tokens since the
last authored boundary, one line per arm, with the marked arm drawn dashed
because it is predicted to lie under the flat arm. Right: the same data as
paired contrasts. The pure frame costs a steep, local charge over the first
few hundred tokens after a deleted boundary and converges to the flat arm
thereafter; the marked-against-flat overlay is E1's notation zero
re-measured as a curve. The axis is nats per byte; the text quotes the same
quantities in front~II's millinats per byte where the two are compared.
Cluster bootstrap by segment over 141 segments,
10,000 resamples, seed 7.}
\label{fig:curve}
\end{figure}

The deficit is local, bounded and steep (Fig.~\ref{fig:curve}). Paired
against the flat arm, the pure frame costs $+0.0335$ nats per byte
[$+0.0245$, $+0.0423$] over the first sixteen tokens after a boundary,
$+0.0085$ over the next forty-eight, $+0.0030$ from 65 to 256, $+0.0009$
from 257 to 1,024, and nothing distinguishable from zero at 1,025 tokens and
beyond. Its scope belongs in the same breath as its result: one reader, one
book, 141 chapter-like segments. It generalizes across distance, not across
works, registers or readers. The pre-registered expectation was near-bin
elevation with convergence by the 1,025 to 4,096 bin, and it held. Put in one
line: the
announcement is worth about a thousand tokens. Its value is re-orientation
at the resumption of prose and not a debt that accumulates across a chapter,
which is the property an operator meant to raise inference demand without
degrading long-run reading has to have.

The units need a sentence, because the two experiments are quoted in
different ones and the ratio between them is a thousand. E6's figures above
are nats per byte, which is what the figure is drawn in; the near-bin deficit
in front~II's millinats per byte is $+33.5$. Set against E1's $+0.47$ that is
not a contradiction and not an error of scale, because the two average over
different things. E1 averages a target window of roughly 1,500 bytes
beginning at a boundary-relative anchor, and E6 averages the first sixteen
tokens after every boundary in one book. The deficit is concentrated at the
resumption of prose and gone by roughly a thousand tokens, so an average over
a long target is small and an average over the first sixteen tokens is not.

Two results from the same run are reported against us. The expectation that
marked and flat are indistinguishable at every bin is violated in the
letter: three late bins have intervals excluding zero, at $-0.00025$,
$-0.00042$ and $-0.00032$ nats per byte, roughly one percent of the
near-boundary pure-frame deficit measured in the same run, though of the
same order as the probe's anchor-level announcement deficit at this reader,
which averages over a much longer window. The pre-registered interior-token
sensitivity column identifies what they are, and the methodological point
generalizes. A full-book pass scores every token, including the
paragraph-final tokens whose byte span carries the separator that follows,
and the three arms write different separators at the 141 pre-announcement
paragraphs; restricted to interior tokens the same bins read $+0.00003$,
$+0.00012$ and $+0.00011$. An anchor design cannot see this effect at all,
because it scores selected targets rather than whole streams, which is why
the anchor-level zero and the full-book violation are consistent rather than
in tension. The third expectation, that the deficit does not grow at long
distance, is reported \texttt{INDETERMINATE}: only sixteen of this book's
chapters run past 4,096 tokens, so the last bin's interval is wider than the
quantity it is meant to bound. One book cannot answer that, which is an
argument for running the curve over the whole pack rather than for reading
the last bin.

One caveat closes the front. Olmo-3 is a hybrid-attention model,
twenty-four sliding-window layers of width 4,096 and eight full-attention
layers, with YaRN engaging above 8,192, so at 7,936 tokens of context only a
quarter of its layers see the whole window. A null from that reader on the
information-gain contrasts is therefore weaker evidence than a null from a
full-attention reader, while its positive announcement-removal result is
exactly as strong. The Qwen3 ladder carries the scale argument and Olmo-3
carries the pipeline-generality argument. The front's five stated limits are
consolidated in \S\ref{sec:epistemic}.

\paragraph{Replication extension (post hoc).} Two extensions were run
during final review, registered as amendments before any GPU pass and
committed to report regardless of direction (A2 to the probe protocol; B1 to
the curve's). A sixth reader from a third training pipeline, SmolLM3-3B-Base,
replicates the announcement deficit on the identical anchor set: $+0.370$
[$+0.040$, $+0.758$] millinats per byte, seed-robust, with all 11 null
controls exactly zero, so the deficit is CI-positive at six of six readers
across three pipelines that share no data decisions. The same reader holds
the swap null on reading behaviour, $+0.104$ [$-0.030$, $+0.255$] at 1.40\%
tightness, and fails the registration's auxiliary condition: its notation
$\Delta$NLL, $-0.152$ [$-0.304$, $-0.025$], excludes zero, a static
marked-context penalty near 0.8\% of mean information gain that no primary
reader showed. On this pipeline the use of long-range context is
notation-invariant while absolute prediction is not, which sharpens rather
than weakens the relocation: the cue that moves reading with distance is the
announcement, and it is one more reason the intervention runs through
announcements rather than sigils. The five-reader primary verdicts are
unchanged. The whole-book curve replicates on a second work, Galton's 1883
\emph{Inquiries into Human Faculty}, serialized by the same operators under
its own pinned structural contract: the pure$-$flat deficit is $+0.0735$
[$+0.0574$, $+0.0906$] nats per byte in the nearest bin, about twice
Moby-Dick's, decays monotonically, and reaches the null inside a thousand
tokens; the marked$-$flat separator artifact recurs at a single far bin with
the same sign and the same interior-only reversal; and the far-tail rule
returns INDETERMINATE by its pre-fixed segment floor, recorded in the
amendment before the run.

\section{Front III: what writers impose}
\label{sec:front-writers}

Front~II asked what a reader uses. This front asks what a writer puts back,
which is the mechanism by which a corpus property becomes a model property,
and the mechanism our own prior asserted without measuring
\citep{freeburg_2026}. The design is front~II's, turned around: the same
works, the same three serializations and the same 7,936-token contexts, with
four base models continuing each context under greedy decoding and the first
1,500 bytes scored by the same detectors the census uses, for markup emitted
per thousand tokens, a boundary-insertion rate and an announcement rate. The
reference level is not zero but the author's own continuation of the same
anchor, scored identically, because authored text carries markup too
(\S\ref{sec:nonull}). \emph{Interior} anchors have
no boundary inside the measurement window, so the three serializations give
byte-identical authored continuations and the human reference is exactly
zero difference; \emph{pre-boundary} anchors are ones where the author's
next move is a boundary, and their pure-label subset carries the question
the format needs answered. Only base models are used, because an
instruction-tuned model's imposition is a post-training artifact and answers
a different question. Rules I1 to I6 were registered before the run. One
count moved between registration and build: the registration's dry run
recorded 29 pure-label pre-boundary anchors, and the deterministic
derivation at build time reproducibly finds 31 under the same label rule.
The delta is documented in the anchor builder and the preflight log, the
human baseline was recomputed on the realized set before any generation
existed, and I3's human reference is exactly zero by construction at either
count.

\textbf{The first-order finding is that they mostly stop writing.} Greedy
decoding on this material degenerates into loops, and that conditions
everything below. The pre-registered guard flags a continuation whose
four-gram repetition rate reaches 0.35, seven times the worst of 483
authored continuations of the same anchors, so it cannot fire on well-formed
prose. It fires on 84\% to 100\% of continuations in every one of the
thirty-six cells: 77 of 1,328 generations survive, three at 0.6B, twelve at
1.7B, twenty-two at 4B and forty at Olmo-3, and mean four-gram repetition
runs 0.84, 0.78, 0.73 and 0.68 up the ladder against 0.005 for the authors.
None of this is a defect of the pipeline. Maximization-based decoding has
been known to degenerate since \citet{holtzman_2020_degeneration}, it is
documented for base models under greedy decoding at billions of parameters
\citep{carlsson_2025_hyperfitting}, and the mechanism is self-reinforcing,
because a sentence already in the context raises the probability of
repeating it \citep{xu_2022_break_loop}. It is the default regime for this
configuration, and the consequence is that this front measures what base
models do in that regime. Far past the protocol's 20\% threshold the guard
stops being a nuisance filter and becomes a selection effect, so every
quantity is reported twice: the exclusion-respecting cells hold two to eight
anchors and cannot carry an interval, so the levels below are the
exclusion-free view over all 69 interior and 41 pure-frame anchors per
model, while the registered verdicts are the exclusion-respecting ones. What
protects the contrasts is that the collapse is common-mode, at similar rates
in all three arms of every model, and the arms stay exactly paired on the
same anchors. The finding still deserves its own line, and the secondary
decoding arm says how to read it. Greedy decoding on this material collapses;
sampling does not. At temperature 1.0 the guard trips on none of the 153 slots
at either reader it was specified for, four-gram repetition falls to between
0.0035 and 0.0059 against 0.0035 to 0.0047 for the authors themselves, and
paragraph lengths return to between 249 and 466 characters against the
authors' 267 to 351, where greedy emits a single unbroken block of roughly
1,400. The collapse is a property of the decoding regime rather than of the
models, which is what the hyperfitting result would predict
\citep{carlsson_2025_hyperfitting}. What is not settled by that is whether
sampled long output is \emph{organized} the way authored text is organized,
and that is exactly what the training-time test of \S\ref{sec:program}
measures.

One reader behaves differently at the margin: Olmo-3 occasionally ends the
document instead of looping, at one anchor of 61 in the flat arm and one of
31 in the pure frame, and no other reader emits an end-of-text token
anywhere. The nearest published observation about generation under a hard
constraint reports formatting richness collapsing rather than rerouting
\citep{one_token_away_2026}, which is about instruction-tuned helpfulness
rather than base continuation, so we note it and do not lean on it.

\textbf{I1, writers impose the register: not satisfied, and the null is
bounded.} On the interior stratum the author opens a new paragraph-level
boundary in 14.5\% of windows and emits no Markdown at all. Handed the flat
serialization, the readers insert boundaries at 5.8\%, 8.7\%, 11.6\% and
15.9\%, none distinguishable from the author's rate except the smallest,
which is below it. Markup emission from a flat context is $+3.53$ tokens per
thousand [$0.00$, $+11.09$] at 1.7B, $+3.53$ [$0.00$, $+11.06$] at 4B,
$+1.82$ [$0.00$, $+5.70$] at Olmo-3 and exactly zero at 0.6B, every interval
touching zero. Stated in the form the registration requires: imposition
above the author, if present at all, is below 11.1 detector-visible markup
tokens per 1,000 generated from a flat context and below 0.15 in
boundary-insertion rate at 95\% confidence, on the widest of the four
readers; from a marked context the 4B's markup excess is distinguishable
from the author's, reported with the exploratory contrasts below. That is a loose bound and it
is the one the data support. Base models handed unmarked prose do not write
the marked register back onto it, which locates the imposition risk in
post-training rather than in base pre-training, and is what the strongest
published counter-evidence to this paper says as well
(\S\ref{sec:epistemic}).

\textbf{I3, the register is restored at silent boundaries: not satisfied,
and this is the front's robust result.} The registered quantity is the
announcement rate in continuations of pure-frame anchors whose deleted
boundary was a pure label. The null is exact rather than estimated: the
serialization removed the announcement, so the authored continuation
contains none and anything the model writes is a reintroduction with no
source in its input. All four readers return 0.000 with a bootstrap interval
of [0.000, 0.000] on the kept set; on the full set three of four are still
exactly zero, the exception being one announcement in 31 continuations at
0.6B, inside a generation whose four-gram repetition rate was 0.941 and
which the guard had already flagged. The human reference is exactly 0.000 by
construction. Set beside front~II's finding that no reader recovers a
deleted announcement from long-range context either, these are two
independent measurements of the same absence, one on the reading side and
one on the writing side. In the terms of \S\ref{sec:pureframe}, a base model
tolerates the pure frame; it does not scaffold it.

That reading needed a discount and now has an answer. A model that has fallen
into a loop reintroduces no
announcement because it is producing little that is structured at all, so an
exact zero measured in that regime is weaker evidence of tolerance than the
number looks. The pre-registration named a temperature-1.0 secondary arm as
the answer to ``greedy suppresses rare events'', and it has now run at the two
readers it was specified for. Handed the pure frame and free to sample, the
1.7B puts no announcement back in 31 continuations, 0.000 [0.000, 0.000];
the 0.6B produces one in 31, a fluent continuation of Arnold whose numbered
endnote apparatus the detector reads as an announcement, the same register
that undermined the reliability audit's precision (\S\ref{sec:invariance}),
for 0.032 [0.000, 0.107], an interval that includes zero, against a human
reference of exactly 0.000. The null holds essentially unchanged where the
models generate fluently, which makes I3 a robust null rather than a
discounted one. We report the secondary beside the primary rather than beneath
it, because the primary's exclusion-respecting I3 cells hold one to five
surviving anchors per reader and the 0.6B's interior marked and flat cells
survive the guard not at all, while the secondary reaches the same conclusion on the full 31
with zero trips. It stays secondary all the same: the registration fixes
greedy as the estimand, because the claim is about the model's default and the
default is the mode of the conditional.

The guard itself then needs a limitation of its own. It was registered as a
nuisance filter, calibrated at seven times the worst authored continuation,
and under greedy decoding it removed 1,251 of 1,328 rows, which is a selection
effect rather than a filter and is why every greedy rate above is the
exclusion-free view. Under sampling it removes nothing at all. A screen whose
behaviour moves that far with the decoding regime cannot be treated as a
property of the text it screens, and a future run of this protocol should
either report both regimes as this one does or drop the guard and report the
repetition statistic beside the rate it conditions.

One exploratory reading of the secondary arm belongs here, labelled as
exploratory and nothing more. The pre-registered free instrument is the
expected number of markup lines along the realized path, read off the
distribution at every line-start step, and it sees an event that no
realization has to contain. Under sampling it separates the arms: marked
context against silent context is $+0.205$ [$+0.107$, $+0.343$] expected
sigils at 0.6B and $+0.212$ [$+0.143$, $+0.292$] at 1.7B, while flat against
silent excludes zero at neither reader. Notation in the input raises the
probability mass the model puts on emitting markup, and the announcement rate
it actually realizes at a silent boundary stays at zero. Input-dependence is
real in the distribution and absent in the realization. We report that as a
distribution-level observation, not as the confirmatory propagation claim that
I2's single-anchor cells cannot support.

\textbf{I2, imposition is input-dependent: the rule returns satisfied and we
report why that should not be read as a finding.} Every cell that triggers
it holds a single anchor from a single work, so its cluster bootstrap
resamples one cluster on every draw, the interval has zero width, and it
excludes zero by construction rather than by evidence. Three of its four
triggering contrasts are negative, while the branch text the rule selected
is the positive one. We therefore treat I2 as unevaluable at these scales
under greedy decoding, which means none of I1, I2 or I3 is satisfied and the
front returns the pre-registered bounded null of I5: the variable is carried
by corpora and used by readers, and base readers at these scales do not
write it back. The experiment that would test the post-training prediction
that follows is the same protocol against an instruction-tuned model of
matched scale, which is an afternoon of GPU.

One pattern in the exclusion-free view is worth recording as exploratory,
and only as that. On the interior stratum, markup emitted from a marked
context is at or above markup emitted from a flat context on every reader,
3.87 against 3.53, 4.07 against 3.53, 1.89 against 1.82 and zero against
zero, while markup from a pure-frame context is exactly 0.000 on all four;
the one contrast against the author that excludes zero is the 4B model's
marked context, at $+4.07$ per thousand [$+0.19$, $+11.65$]. Should that
ordering survive a decoding regime that does not collapse, it would say
notation in the context is carried into the continuation, and the general
form of such an effect is documented and localized to two to four percent of
attention heads across 26 models \citep{entrainment_2026}. We do not claim
it here, and the same work reports entrainment \emph{decreasing} with model
size, which cuts against reading our own I4 as scale stability.

\textbf{I4, imposition scales: not satisfied, and indeterminate rather than
flat.} The registered ladder quantity is 0.500 [0.000, 1.000] at 1.7B and
0.250 [$-0.500$, 1.000] at 4B, with no value at 0.6B, because the
exclusion-respecting interior cells are where the collapse bites hardest.
Those cells cannot support a scaling claim in either direction, so we report
the rule as unsatisfied and the question as open rather than as evidence of
scale stability.

\textbf{I6, the determinism gate: passes, and how it came to pass is the
point.} Eleven null-control anchors per model have byte-identical prompts in
two arms and must therefore produce byte-identical text; all four readers
return 11 of 11. On the first attempt they did not, and the cause was found
rather than guessed. The generation library admits sequences into a prompt
batch in sub-groups, so two arms of one anchor falling on opposite sides of
a sub-group boundary begin decoding at different steps and thereafter see
different batch shapes; different shapes select different kernels and
reduction orders, a near-tie in the argmax flips, and greedy decoding
amplifies one flipped token into a whole different continuation. The two
offenders diverged at generated characters 46 and 54, and the third pair,
whose arms fell inside one sub-group, was byte-identical as predicted. The
recorded amendment, A1 in the log, prefills each batch as a single group,
which makes two
byte-identical prompts two rows of the same matrix multiplication at every
step and so bitwise identical by construction rather than by luck. It was
adopted mid-run and is disclosed here rather than left in the results
directory: it changed batch scheduling only, the estimand and the decision
rules were untouched, it also cut measured peak memory on the reader it was
calibrated against from 28.75 to 8.45 GB, and the
pre-amendment run directory was deleted so no row produced under the old
schedule is used anywhere above. A gate that fails, finds its cause and then
holds is better evidence about a pipeline than one that never fails. The
secondary arm produced a second defect of the same species and it is disclosed
on the same terms: the exclusion gate written for greedy was still being
applied under sampling, so rows that had tripped nothing were being
quarantined, and the guard markers are now written per arm, so a guard result
carries the arm it was computed in. No number above was produced under the
defective quarantine.

\begin{table}[t]
\centering
\footnotesize
\setlength{\tabcolsep}{5pt}
\begin{tabular}{lrrrrrlr}
\toprule
 & & \multicolumn{3}{c}{markup / 1k} & boundary & announcements & rep.\ \\
\cmidrule(lr){3-5}
reader & params & marked & flat & pure & rate & restored (pure) & trips \\
\midrule
Qwen3-0.6B-Base & 0.60B & 0.00 & 0.00 & 0.00 & 0.058 & 0.000 \ci{0.000}{0.000} & 69/69 \\
Qwen3-1.7B-Base & 1.72B & 3.87 & 3.53 & 0.00 & 0.087 & 0.000 \ci{0.000}{0.000} & 67/69 \\
Qwen3-4B-Base   & 4.02B & 4.07 & 3.53 & 0.00 & 0.116 & 0.000 \ci{0.000}{0.000} & 65/69 \\
Olmo-3-1025-7B  & 7.30B & 1.89 & 1.82 & 0.00 & 0.159 & 0.000 \ci{0.000}{0.000} & 61/69 \\
\midrule
\textbf{the author} & --- & \textbf{0.00} & \textbf{0.00} & \textbf{0.00} & \textbf{0.145} & \textbf{0.000} (by construction) & \textbf{0/69} \\
\bottomrule
\end{tabular}
\caption{Front III on the interior stratum, where the three serializations
produce byte-identical authored continuations and the author's reference is
therefore exactly zero difference. Markup is detector-visible tokens per
1,000 generated, the mean of per-continuation rates (pooled rates are
within 8\% everywhere), by the notation of the visible context; boundary rate is
the fraction of continuations opening a new paragraph-level boundary, from
the flat context; \emph{announcements restored} is the pre-registered I3
quantity, the rate at which a reader writes an announcement into a pure-frame
continuation whose own boundary was deleted, over 31 pure-label anchors in 12
works. Decoding is greedy at 384 generated tokens from 7,936-token contexts
with the end-of-text token unsuppressed, so there is no sampling seed;
greedy was pre-registered because it is the estimand, what a base model does
when it maximizes. The temperature-1.0 companion registered as the answer
to ``greedy suppresses rare events'' has run at 0.6B and 1.7B on the
pre-boundary stratum and is reported in the text as a secondary arm.
Bootstrap intervals
are clustered by work, 10,000 resamples, seed 7. \emph{Rep.\ trips} is the
pre-registered repetition guard firing on the interior flat arm, and it is
the reason the levels in this table are the exclusion-free view over all
anchors while the verdicts in the text are the exclusion-respecting ones:
77 of 1,328 continuations survive the guard across the four readers, three
at 0.6B and forty at Olmo-3, leaving exclusion-respecting interior cells of
one to five anchors in the \emph{announcements restored} column and bounding
nothing. Those intervals are zero-width because they resample one cluster, so
they are degenerate rather than tight; the temperature-1.0 secondary arm
returns 0 of 31 at 1.7B and 1 of 31 at 0.6B with no trips at all, and the
text reports both beside this table for that reason. The
guard never fires on the author's own prose. Four readers rather than
five: the protocol's arbitration rule gave the machine to this front rather
than to the 8B probe row, which ran afterwards and appears in
Table~\ref{tab:crossscale} with no generation arm to match it, so the
same-family ladder here is three
points and the reduced strength is accepted rather than papered over.}
\label{tab:imposition-gen}
\end{table}

\begin{figure}[t]
\centering
\includegraphics[width=\linewidth]{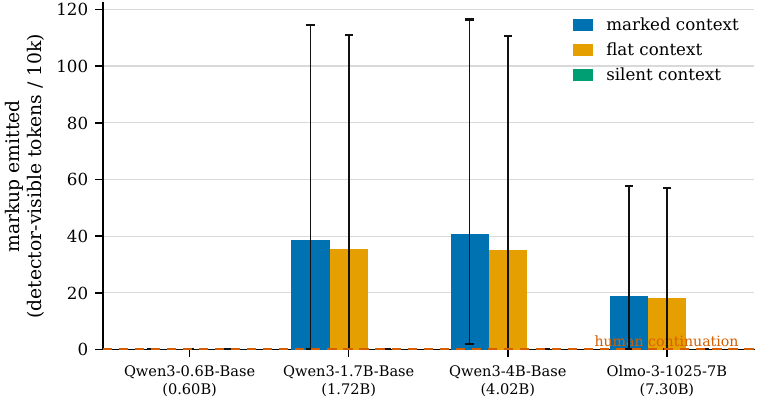}
\caption{What a base model writes back, by the notation it was shown.
Detector-visible markup per 10,000 generated tokens on the interior
stratum, with 95\% cluster-bootstrap intervals and the authored continuation
drawn as the reference line, which sits at zero because interior authored
text carries no markup. Every interval reaches zero except the 4B model's
marked-context bar, which is the one cell in which a base reader's markup
emission is distinguishable from the author's. The pure-frame bars are
exactly zero on all four readers.}
\label{fig:imposition}
\end{figure}

\section{The pure frame}
\label{sec:pureframe}

Three measurements in this paper compose into a design, and the design is
narrower than the one we set out to make. Flattening a work's notation while
keeping its heading words leaves the boundaries detectable in the same places
(\S\ref{sec:invariance}). Swapping the notation on identical text moves a
trained reader's use of long-range context by a measured zero, at every
reader scale we have run (\S\ref{sec:front-readers}). Deleting the
announcement itself is the only manipulation that changes reading behaviour
at all. Together those say something uncomfortable about the obvious remedy:
\textbf{faithful flattening is a no-op}. Stripping Markdown from a novel and
leaving \texttt{CHAPTER IV.} standing on its own line produces a file that is
different in notation and identical in what it demands. The sigil axis is a
distraction, and it is our own zero that retires it. The pre-registered
qualifier belongs in this paragraph rather than in a footnote, because the
whole-book curve of \S\ref{sec:front-readers} violated the notation null in
the letter at three late distance bins. Those violations are roughly one
percent of the curve's near-boundary deficit and they are separator-token
effects at paragraph edges, not effects on prose: restricted to interior tokens the
same bins are positive and negligible. So the no-op claim is scoped to
interior prose, which is what the sentence is about, and is not a claim that
the two files are identical token for token.

What is left is the announcement: a short line, set off by whitespace, whose
function is to say that a boundary is here. The material that demands the
inference is the material with no announcements in it at all. We call that
the \textbf{pure frame}, and we specify it rather than gesture at it.

\begin{itemize}[leftmargin=1.4em,itemsep=2pt,topsep=3pt]
\item \textbf{Keeps} body paragraphs, in authored order, separated by exactly
  one blank line. Sentence-level punctuation and intra-paragraph typography
  are untouched.
\item \textbf{Deletes} every structural announcement line: notation sigils
  and the lines carrying them, pure labels (\texttt{CHAPTER IV.},
  \texttt{PART ONE}), titled heads (\texttt{Chapter 3: The Meeting}), running
  heads, page furniture, transcription artifacts and footnote-marker lines.
  Operationally, a line is an announcement if the census detector of
  \S\ref{sec:instrument} fires on it, or if it is notation syntax.
\item \textbf{Renders the boundary as nothing.} A deleted announcement leaves
  adjacent paragraphs separated by the same single blank line as any other
  two paragraphs. A double gap would itself be an announcement.
\item \textbf{Records every deletion} in a per-document sidecar: the deleted
  text, its class, its character offset in the source serialization, and the
  paragraph index it preceded. The transform is reversible, and the audit is
  the artifact's warranty.
\end{itemize}

A file passes the validator if and only if it contains zero announcements and
every record's sidecar round-trips byte-exactly. The wording matters, because
\S\ref{sec:instrument} showed that the detector's recall on titled heads is
2.8\% on a novel. The detector is a \emph{screen}, not the definition; the
ground-truth announcement list is the definition, and the sidecar is the
warranty. Where those agree, $S(W) = 1.0$ for every $W$ up to document
length, and the validator is what turns that from a claim about a pipeline
into a checkable property of a shipped file.

Building it surfaced eleven questions the specification did not answer, and
three of them are worth a sentence each because anyone implementing this
will meet them. Deletion is \emph{partial} where only a line-initial sigil
fires: a 300-word paragraph its author happened to number keeps its words and
loses its \texttt{32.\ } prefix, which is the spec's blockquote clause read
at prefix granularity. Deletion \emph{iterates to a fixed point}, because
removing one line can leave the line beneath it newly standalone and
therefore newly announcing; the pass number is recorded per edit. And the
sidecar is an ordered list of non-overlapping \emph{replacements} over the
flat text rather than a list of deletions, which is what makes the inverse
well defined and the round trip exact in both directions.

Three honesty clauses travel with the spec. Deleting titled heads deletes
authored words, which no amount of framing makes free; the sidecar makes it
reversible, and in the recommended mixture the marked copy coexists, so the
corpus loses nothing and only the pure \emph{copy} does. Paragraph breaks
remain, and the reason is exactly the mechanism of this paper: a paragraph
break fires at every paragraph, so its reliability as a discourse-boundary
cue is close to zero, and a cue with no reliability is not a shortcut.
Furniture classes also differ by material. Version 0.1 specifies literary and
argument prose; proceedings, correspondence and technical writing each need a
class list before the spec applies to them.

We ship two packs over the same 27 public-domain works, because the
comparison is the argument. The \emph{flat} pack is the faithful baseline:
2,313,460 tokens, one record per book, notation removed and heading words
kept verbatim. Its markup claim has to be stated carefully, because the
builder's original guard was a tautology that could not fire. What is true is
that the flattener writes no Markdown of its own: no \texttt{\#}, no rule, no
bullet glyph, no case change. Under this paper's own census detector, nine of
the twenty-seven records still light up, for 826 tokens in 280 spans, or
0.036\% of the pack, and the cause is inspected and benign: 272 of those
spans are numbered chapter headings whose author's own text begins with a
numeral, as in \texttt{2. Mistakes about Political Economy}, six are
\texttt{>From} line-starts inherited from a Gutenberg source file, and two
are footnote markers. The flat pack's $S(8{,}192)$ is therefore 0.836,
against 0.153 for the olmOCR slice and 0.889 for C4, with the caveat of
\S\ref{sec:instrument}: C4's figure is computed over the small tail of its
documents that reach one window, and the pack's over documents that are all
far longer than one. It is a better number
for this paper than the guarantee it replaces, because a reviewer can check
it.

The \emph{pure} pack is the format itself. The same 27 works come to
2,297,423 tokens after 1,577 lines and 16,502 tokens are deleted (the
pure total sits 465 tokens above flat minus deletions: a deletion junction
breaks byte-pair merges and the remainder retokenizes), which is
0.713\% of the flat pack. Three survival results come out of it, two of them
warranties and one of them a tautology, and it is worth separating them.
Both census detectors firing zero times over the result is a fixed point of
the deletion rule rather than independent evidence: the builder deletes
whatever the screen fires on until the screen stops firing, which is the same
objection this paper just made to the flat pack's original guard. What is
independent is the ground truth, the announcement list the intermediate
representation emitted, against which $S(8{,}192) = 1.000$ over
267 windows and $S(32{,}768) = 1.000$ over 57, and in all 27 records the
longest markup-free run is the entire work; and the sidecar, which
round-trips byte-exactly in all 27 records. The detector is the screen; the
sidecar is the warranty. The cost is concentrated exactly where the honesty
clause says it will be: titled heads are 10,257 of the 16,502 deleted tokens,
62\% of everything removed, while pure labels cost 1,118 tokens across the
whole pack. Deleting \texttt{CHAPTER IV.} is free and deleting \emph{Of the
Limits to the Authority of Society over the Individual} is not.

Two further disclosures belong with the artifact. The fixed-point rule
deletes whatever the screen fires on, and on this pack that includes 319
short standalone lines the manifests type as body text: signature blocks,
one-line dialogue turns, an epigraph fragment, and in the mixed-material
works a great many table rows and index entries. They are listed individually
in the artifact rather than summarized, because they are the operator's audit
surface. And the format is not a compression scheme: on \emph{Moby-Dick} the
deletion buys 1,587 tokens of headroom, the pure stream is still 283,267
tokens, and it still does not fit a 256k context. The whole gain is in what
the stream demands.

The recommendation is a mixture, and this paper's own data settles which axis
it runs on. It is not a mixture over notation, of Markdown against plain
text, because the notation swap is a measured zero. It is a mixture over
\emph{announcement presence}: the same work present both marked and pure, so
that at the same underlying event the announcement is sometimes there and
sometimes not. That destroys the cue's reliability at the level where
reliability does its work, and destroys nothing else, because the marked
copies remain, structural competence is still taught, and every copy is the
same author's words. Aim it at the long-context stage of \S\ref{sec:slot}.

We state the central hypothesis as a bet, with its price. We are betting that
the factorization exists and that reliability is what forces it: that a model
trained on the same arrangement with and without its announcements forms a
discourse channel separable from the surface channel that renders it. If the
paired arm of the three-arm run (\S\ref{sec:program}, test 3) is
indistinguishable from the marked arm, reliability was not the operative
variable, the factorization story collapses to a notation preference, and we
will report it as such.

One last thing about what the operator is for, because it is the part most
easily mistaken for archival fussiness. Fidelity is the converters'
objective, and \S\ref{sec:converter} shows that they achieve it. The pure
frame has a different target: it is a repeatable operator that turns a work
into what it already is, an unbroken stretch of sustained thought, which is
the material pre-training is not currently producing. It is judged by the
output it produces, which the
validator checks, rather than by fidelity to its input, so edge cases in
furniture classification need to be repeatable and audited rather than
archivally perfect. The pure frame is a format, not an edition. Nobody reads
it; models train on it. The general form of that point is the sentence we
would most like carried out of this paper: \textbf{choose format operators by
the capability they train, not by the fidelity they preserve.} And if the
objection is that books are 0.2\% of a modern mixture, smallness is the
economics of the proposal rather than an embarrassment. A slice carrying a
disproportionate share of a corpus's organizational variety is exactly what a
volume-keyed accounting system cannot see.

\begin{table}[t]
\centering
\small
\begin{tabular}{lrr}
\toprule
quantity & flat pack (baseline) & pure pack (the format) \\
\midrule
works                                & 27 & 27 \\
tokens (\texttt{o200k\_base})        & 2,313,460 & 2,297,423 \\
median work                          & 73,724 & 73,602 \\
\addlinespace
fits in a 32,768-token window        & 0 / 27  & 0 / 27 \\
fits in a 65,536-token window        & 12 / 27 & 12 / 27 \\
fits in a 131,072-token window       & 23 / 27 & 23 / 27 \\
fits in a 262,144-token window       & 27 / 27 & 27 / 27 \\
\addlinespace
fragments when packed at 8,192       & 297 & 297 \\
authored markers per fragment        & 4.13 pooled; 3.86 unweighted & 0 \\
\addlinespace
lines deleted                        & 0 & 1,577 \\
tokens deleted                       & 0 & 16,502 (0.713\%) \\
\quad of which titled heads          & --- & 10,257 (62\%) \\
\quad of which pure labels           & --- & 1,118 \\
notation-detector markup             & 826 tokens, 280 spans, 9 works & 0 \\
typographic-detector lines           & 1,480 & 0 \\
pack-wide \Sw{8{,}192}               & 0.836 & 1.000 (267 / 267 windows) \\
pack-wide \Sw{32{,}768}              & --- & 1.000 (57 / 57 windows) \\
sidecar round-trips, byte-exact      & --- & 27 / 27 \\
\bottomrule
\end{tabular}
\caption{The two long-form packs over the same 27 public-domain works. The
flat pack is the faithful baseline: notation removed, heading words kept
verbatim. Its detector-visible markup is not imposed structure but kept
authored text, 272 of the 280 spans being numbered chapter headings whose
author's own words begin with a numeral; the flattener writes no Markdown of
its own. At $S(8{,}192) = 0.836$ it stands against 0.153 for the olmOCR slice
of Dolma~3 and 0.889 for C4, subject to \S\ref{sec:instrument}'s caveat that
the web figure is computed over the tail of C4 that reaches one window while
the pack's is computed over documents that all exceed it many times over. The
pure pack deletes the announcements and
records every deletion in a sidecar. Its detector-zero row is a fixed point
of the deletion rule and not evidence; the independent warranties are the
ground-truth survival column and the 27 byte-exact sidecar round trips.
Fragment counts are per document
at an 8,192-token packing length and are a property of the works rather than
of the serialization, which is the point of printing them twice.}
\label{tab:pack}
\end{table}

\section{Three settling tests}
\label{sec:program}

The program this paper belongs to has three tests, in increasing order of
cost. One has been run, one is piloted here, and one is specified and costed
and has not been run.

\textbf{Test 1, the fabrication test. Run; it returned a null against us.}
The prediction was that current converters would fabricate structure on
prose, and that the damage would concentrate on exactly the material still
carrying organizational variety, because heading detection is a classifier
and continuous prose is the class on which its positives should mostly be
false. It needed no model training and a few days of compute. It is reported
in \S\ref{sec:converter}, and it failed: imposition loss is significantly
negative on every prose stratum, so the converters under-mark rather than
invent. The fabrication mechanism is dead. What replaced it is the
reliability mechanism, which is stronger, because a faithful cue is a
trustworthy one and trustworthiness is what a shortcut needs.

\textbf{Test 2, the reader-side test. Piloted here.} The original form of
this test compared public checkpoints from families whose pipelines are
documented to differ in structural handling, and looked for benchmark scores
and long-form prose likelihood to separate. Front~II is that test in a
sharper form, because it holds the content byte-identical across arms instead
of comparing families whose corpora differ in a hundred ways at once, and
because it runs the same anchor set across a ladder of reader scales. What it
can settle is whether trained readers use the announcement, and whether the
notation carrying the announcement matters. What it cannot settle is why: a
reading-side probe cannot distinguish ``the training pressure never existed''
from ``these particular readers never formed the machinery''. Front~III
narrows that gap from the writing side by asking whether models put the
announcements back, and they do not. Only test 3 closes it.

\textbf{Test 3, the decisive test. Specified and costed; not run.} It needs
three arms on identical content rather than two. The obvious two-arm version,
plain text against Markdown, does not work, because it conflates faithful
re-serialization with imposed hierarchy and therefore cannot separate this
paper's claim from the trivial claim that notation matters. An earlier
design's third arm was imposed structure, a detector running over continuous
prose and emitting hierarchy it inferred; our own notation zero retires that
axis, and the protocol records the retirement. The three arms as specified
are \emph{marked}, current practice, announcement reliability near 1.0;
\emph{pure}, the format of \S\ref{sec:pureframe}, reliability zero; and
\emph{paired}, half the works marked and half pure, partitioned by work at
matched tokens rather than duplicated, so the manipulation is not confounded
with repetition, taking the stream's announcement reliability to one half.
Every word, token budget and data order is identical across arms; only the
serialization moves.

The cost matters, because a costed experiment is an invitation and an
uncosted one is a wish, so the costing ships as a repo artifact with its
arithmetic exposed rather than as a number in a sentence. At one billion
parameters and 100 billion tokens per arm, the three-arm run is an estimated
3,100 H100-hours, which is 1.4\% of one OLMo-3-7B pre-training run by the
same arithmetic, or about two days on 64 accelerators; a screening version at
29 billion tokens per arm is 0.7\%. The cheaper precursor is cheaper still
and costs no tokens at all, because it is a substitution rather than an
addition: replacing 5 to 25 billion tokens of an Olmo-3-style 50-billion-token
long-context stage with pure-frame material adds one branch to a stage whose
infrastructure is already configured. Those are derived estimates from
published throughput and token counts, not measurements, and the protocol
states them that way. The readouts are the instruments this paper ships, run
per arm over held-out works: the probe's announcement contrast, the
whole-book curve and the generation probe, with the loop-collapse rate
reported as a first-class outcome because it is the number that connects the
variable to a product surface. They run beside the laboratory's own suites,
under decision rules pre-registered in the style of R1 to R4 and I1 to I6.
We pre-commit the direction in both cases. If the paired arm is
indistinguishable from the marked arm, reliability was not the operative
variable and the factorization of \S\ref{sec:pureframe} collapses to a
notation preference; and if the slot substitution moves nothing on any
instrument, the locus argument fails, cheaply, which is the point of running
it first.

\section{Epistemic status and limitations}
\label{sec:epistemic}

The strongest published evidence against this paper's framing is a
base-model result, and it belongs here rather than in a reviewer's report. A
four-way classifier over model families, at 25\% chance, identifies
instruction-tuned models from markdown features alone at 77.7\% and base
models at only 38.5\% \citep{sun_2025_idiosyncrasies}. Base models mostly
emit plain prose, which suggests that much of the visible chunking is
post-training policy rather than pre-training inventory. The same table has
a second row that is rarely quoted and that we think settles the reading: on
the \emph{original} text rather than its markdown features, the same
classifier identifies base models at 87.3\% against 96.3\% for instruct. So
base models are highly identifiable, just not from markdown. Post-training
decides how much of the tendency is \emph{expressed}; what pre-training
supplies is what there is to express. That distinction is a hypothesis, not
a finding, and the honest concession is that post-training demonstrably does
the visible compression: matched-checkpoint work isolates narrative
flattening to post-training directly \citep{narrative_flattening_2026}. The
open question is what range the base inventory had, and nobody has measured
it, because notation was never a coordinate. Our own prior found the latent
tendency present before post-training in a base-versus-instruct comparison
\citep{freeburg_2026}; that is our result and not the idiosyncrasies
literature's, and we would rather say so than let the two blur.

A second counter-current is fresher and cuts the same way. A prefix-removal
probing method recovers 47.7\% of the original records as clean units from
flattened, separator-stripped text, against 25.9\% for the strongest
embedding baseline \citep{right_reset_2026}. That is a single-author preprint, and it is an
\emph{intervention}: 47.7\% is what the probing method can extract, not what
a model does unprompted. Read correctly it is the right correction to make to
this paper's language. Boundaries are degraded, not absent, and the claim
that survives is about how much pressure exists to make the reconstruction
good.

The reader-side front carries five limits that we state whatever its
verdicts turn out to be. It measures reading and not training, so it cannot
distinguish an absent training pressure from readers that never formed the
machinery. Its announcement arm covers twelve of the twenty-one works, for
the reason given in \S\ref{sec:front-readers}, so the scale ladder is a
ladder over that subset. The notation contrast is a tight null rather than a
wide one, which bounds any suppression-style effect below 1.7\% of measured
information gain at every reader scale and at 0.7\% on the published one.
The rise in prediction difficulty
after announcement removal is largest immediately after the boundary in the
point estimates at every reader, but it is not monotone in distance, it does
not replicate as an interval, and it must not be read as decay. The
near-boundary stratum excludes zero only at the published 0.6B reader, at
$+0.615$ [$+0.172$, $+1.143$]; the same stratum returns $+0.238$ [$-0.137$,
$+0.613$] at 1.7B, $+0.410$ [$-0.082$, $+0.927$] at 4B, $+0.775$
[$-0.165$, $+1.641$] at the cross-family reader and $+0.273$ [$-0.116$,
$+0.673$] at 8B, all four covering zero.
The middle and far strata, $+0.175$ and $+0.605$ with the far interval
including zero, were computed at the 0.6B reader only and are not ladder-wide
results. And one anchor, also at that reader, breaks the arm's implicit
null while the result survives it: of six anchors whose long context contains
no boundary, five return exactly zero as the pipeline requires, while the
sixth, the smallest target in the set at 571 bytes, returns $+5.91$, which is
consistent with token-grid drift from the upstream deletions rather than any
boundary effect. Dropping it gives $+0.404$ [$+0.110$, $+0.692$], still
excluding zero, and leaves the near-stratum headline unchanged. Weights are
bf16 with float32 scoring throughout, and the eleven exact-zero control
anchors come from seven of the twenty-one works.

One statistical posture covers the whole paper and is easier to state once
than to defend seventeen times. Every interval printed here is a
pre-registered contrast evaluated once, not a survivor of a search: the
rules, the strata, the horizons and the thresholds were fixed before the runs
that decide them, and where a rule admits several triggers we report how many
were eligible and how many fired, as \S\ref{sec:front-readers} does for R2
and \S\ref{sec:front-writers} does for I2. Two conventions vary and are
labelled where they occur: the census supply rows use 1,000 bootstrap
resamples where everything else uses 10,000, and the writer-side front
reports every rate twice because its exclusion filter crossed the threshold
at which the registration said it stops being a filter. Where a bootstrap
interval has zero width it is a degenerate resample over one or a few
clusters, and it is named as such wherever it is printed rather than being
allowed to read as a tight interval.

The pre-registered rules were evaluated mechanically on all five
readers and reported as they fell (\S\ref{sec:front-readers}). Three went
our way and one did not. The announcement deficit persists at five of five
scales (R1) and the notation swap is a tight null at five of five (R3),
which is what the framing predicts; the determinism gate passes at five of
five (R4), which is a statement about the pipeline. But R2, which asked
whether larger readers reconstruct a deleted announcement from context,
fired at 1.7B, and its branch text was written in advance to say that this
cuts against the paper. It does, and we report it as a headline rather than
a footnote, with the qualification the diagnostics require: the effect
appears at one of the four eligible readers, is absent at all three larger
ones, and turns on a bound that stays above zero at every seed run, by a
margin at its closest seed smaller than the reproducibility envelope fixed
before the run. A reader who wants the strongest form of the
objection should take it as stated and wait for the training-time test,
which is the only instrument that separates ``the pressure never existed''
from ``these readers never formed the machinery''. The fifth reader landed
after the writer-side runs and all four verdicts were recomputed over the
complete ladder; none of them changed direction. A post-hoc sixth reader
from a third pipeline replicates the deficit and holds the reading-side
swap null while failing its auxiliary $\Delta$NLL condition, a small
static notation sensitivity; the replication and the failed conjunct are
reported together in \S\ref{sec:front-readers} (Amendment A2).

The writer-side front carries four limits of its own, and the first is
large. Greedy decoding collapsed into repetition on 84\% to 100\% of
continuations in every cell, leaving 77 of 1,328 generations on the kept
set, which is far past the threshold at which the pre-registered guard
becomes a selection effect rather than a filter; every rate is therefore
reported twice, and the exclusion-respecting cells are small enough that
only the exact zeros should be leaned on. Second, the one rule that returned
satisfied, I2, is carried entirely by single-anchor cells whose bootstrap
has zero width, and three of its four triggering contrasts run the opposite
way to the branch text it selected, so we treat it as unevaluable and report
the front as the pre-registered bounded null (\S\ref{sec:front-writers}).
Third, the scaling rule I4 is indeterminate rather than flat; the secondary
temperature-1.0 arm has run and shows the collapse to be an artifact of greedy
decoding, but it covers two readers and one stratum, and the registration
keeps greedy as the estimand, so the primary's small cells stand as the
primary. Fourth, the front measures four
base models on public-domain literary prose at one context length, which
bounds what its null can mean: it says these readers do not impose this
register on this material, not that no base model imposes any.

One of this paper's own pre-registered tests returned against its own
slogan, and we would rather that be visible here than discovered. The
reliability of a structural sigil as a boundary announcement was registered
with a threshold and measured, and the verdict is \texttt{UNDERMINED} as
stated: the claim holds in narrative at precision 0.994, or 0.967 once the
documented typographic furniture is removed from the ground truth as the
registration required, and fails in sustained argument at 0.765, or 0.606
adjusted, carried there by one document's endnote
apparatus (\S\ref{sec:invariance}). The adjusted narrative figure puts the
claim in the registered SUPPORTED, WEAKENED band rather than the SUPPORTED AS
STATED one, which is why this paper writes near-perfect reliability and never
perfect reliability. The mechanism therefore has a scope, and
the scope happens to be the material the proposal is about, which is
convenient enough that it deserves saying out loud rather than quietly.

Three more limits belong on the record and are argued where they arise. The
census cannot be fully regenerated: seven of its ten published rows rest on a
run that wrote no output file, its survival column rests on few windows at
the re-verification budget, and one row diverges on re-measurement
(\S\ref{sec:census}). The converter study contains no vision-language
converter, and its registered rule was close to unsatisfiable at the
evaluation horizon, which is why the satisfiable companion at $W = 2{,}048$
is reported alongside it (\S\ref{sec:converter}). And the motivation this
work started from remains a motivation: readers report a recognizable
flatness in machine-generated prose, everything outlined and signposted and
closed with a neat turn, and if that texture is real it is what a thin
inventory looks like from outside. It carries a withdrawal condition, because
an aesthetic judgment invoked as evidence needs one: strip the formatting
from prose of both kinds and have raters judge it blind, and if their
judgments are predicted by whether they believe a machine wrote it rather
than by organizational measures computed on the stripped text, the motivation
is an artifact of disclosure and should be withdrawn without prejudice to
anything measured above. That experiment has not been run.

\begin{table}[t]
\centering
\small
\begin{tabular}{p{0.30\linewidth}p{0.34\linewidth}p{0.28\linewidth}}
\toprule
claim & what would falsify it & pre-committed response \\
\midrule
Notation is unrecorded, and $S(W)$ is new
  & a dataset card reporting markup prevalence, or any published statistic of
    the $S(W)$ family
  & withdraw the novelty claim; the measurement stands \\
\addlinespace
Converters fabricate structure on prose
  & \textbf{already falsified} (\S\ref{sec:converter})
  & reported as ours; mechanism replaced by reliability \\
\addlinespace
The announcement is the operative cue, not the sigil
  & a notation swap whose interval excludes zero at any reader scale
  & the mixture axis returns to notation; the pure frame loses its warrant \\
\addlinespace
Readers do not reconstruct a deleted announcement from range
  & \textbf{fired once} at 1.7B, not at 4B, 7.3B or 8.2B
    (\S\ref{sec:front-readers})
  & reported as a headline, with its non-replication and its seed margin \\
\addlinespace
Reliability is the operative variable at training time
  & the paired arm of test 3 indistinguishable from the marked arm
  & the factorization collapses to a notation preference; report as such \\
\addlinespace
The long-context stage is where it binds
  & markup-controlled slot substitution changing nothing
  & the locus argument fails, cheaply \\
\addlinespace
Structural variety is an outcome worth caring about
  & blind raters' judgments predicted by believed authorship rather than by
    organizational measures
  & withdraw the motivation without prejudice to the measurements \\
\bottomrule
\end{tabular}
\caption{The falsifier ledger, with directions committed in advance. Three
entries are already settled: the first row's claim survives a live survey
(\S\ref{sec:crux}), the second row's claim is our own and is dead, and the
fourth row's falsifier fired at one reader of five and is reported as the
pre-committed response requires.}
\label{tab:falsifiers}
\end{table}

\section{Record it: a data card for notation}
\label{sec:datacard}

The smallest useful response to everything above is instrumentation, and
almost all of it already exists. A dataset card that records what a document
passed through on the way in, plus one number computed over tokens the
pipeline already holds, makes the variable visible retrospectively and the
question askable for every corpus built after it. Table~\ref{tab:datacard} is
the proposal. What the card records is the serialization register of
\S\ref{sec:variable}; the authorial composition register it can only
preserve evidence for, in the source that \texttt{extractor} and
\texttt{ocr\_path} point back to, which is why these fields belong at
ingestion rather than in an audit like ours.

The framing matters more than the field list, because it changes what is
being asked. FinePDFs already carries a per-record \texttt{processor} field,
declared in its own schema as determining which PDF extractor produced the
sample \citep{finepdfs_2025}. A major 2025 corpus therefore already records
extractor identity per document. The proposal here is not new
infrastructure. It is to generalize a field the field's own flagship PDF
corpus already has, and to add the one statistic nobody computes. That is a
request to finish something rather than to adopt something, and the two land
very differently.

The reason the statistic is a realistic ask is the reason it looked like a
weakness in \S\ref{sec:instrument}. $S(W)$ is a surface count over a
tokenized stream: deterministic, seedless, and reproducible by anyone with
the corpus and an afternoon. No laboratory is going to run someone else's
RoBERTa over its pre-training mixture and reproduce someone else's annotation
guidelines, and none should have to. Reporting $S(W)$ at the context lengths
the corpus will actually be trained at costs a pass over data the pipeline
has already tokenized once.

The last row of the table is the lesson of our own census, and we would
rather propose it having been caught by it. Document count, token count,
sampling rule, corpus revision and date are the five things that turn a
statistic into a measurement, and they are exactly the five things our own
published rows could not supply for seven of ten corpora. A corpus statistic
without its sampling rule is a number, not a measurement, and the Hansard row
of \S\ref{sec:supply}, whose median document moves by a factor of almost
eight between two sampling rules, is the demonstration.

Documentation tooling has not closed this gap on its own. A 2026 scoping
review of dataset-documentation tools finds no pipeline-level extraction
fields anywhere in the tooling it surveys \citep{doc_tools_review_2026}, and
the automated-quality-metric line that has grown up alongside datasheets has
no notation axis either \citep{datarubrics_2025}. The fields below are cheap
and they are not being asked for by anything currently in use.

\begin{table}[t]
\centering
\small
\begin{tabular}{lp{0.62\linewidth}}
\toprule
field & what it records \\
\midrule
\texttt{extractor}        & tool and version that produced the text from its source
                            (e.g.\ \texttt{trafilatura 2.2.0}, \texttt{resiliparse},
                            \texttt{jusText}, \texttt{Docling}, \texttt{RolmOCR}) \\
\texttt{extractor\_flags} & the arguments that decide notation, named explicitly
                            (\texttt{output\_format}, \texttt{include\_formatting},
                            \texttt{favour\_precision}) \\
\texttt{ocr\_path}        & whether the record passed through a vision-language
                            or OCR stage, and which one \\
\texttt{conversion\_target} & the notation the pipeline emitted: plain text,
                            Markdown, HTML, or a mix \\
\texttt{s\_w}             & clean-window survival at the context lengths the
                            corpus will be trained at, at minimum
                            \Sw{8{,}192} and \Sw{32{,}768}, together with the
                            share of documents long enough to contribute a
                            window at each $W$ \\
\texttt{markup\_per\_10k} & core structural markup tokens per 10{,}000 tokens \\
\texttt{tokenizer}        & the tokenizer the two statistics above were computed under \\
\texttt{sample}           & document count, token count, sampling rule, corpus
                            revision, and date --- the five things the census
                            this paper reports could not recover for seven of
                            its ten rows \\
\bottomrule
\end{tabular}
\caption{Proposed data-card fields for notation. The first four generalize a
field FinePDFs already carries per record. The middle three are the one
statistic nobody computes, plus the tokenizer it was computed under. The last
is the lesson of our own census, where all five of its components were
unrecorded for seven of ten rows.}
\label{tab:datacard}
\end{table}

\paragraph{Artifacts.} The two packs and their sidecars, the census with
per-row provenance (\path{data/census/census-table.json}), every
experimental artifact a printed number in this paper traces to, the claims
ledger that enforces the trace, the pure-frame implementation and validator,
and the costed laboratory protocol of \S\ref{sec:program} are published at
\url{https://github.com/emfreeburg/announcement-carries-cue}.

\section{Related work}
\label{sec:related}

\textbf{Extraction.} Three groups arrived independently at the position that
extraction is an under-examined variable, and all three measured its
consequences rather than its product. AICC builds a 7.3T-token corpus with a
model-based HTML parser and beats trafilatura-extracted text by 1.08
percentage points across thirteen benchmarks \citep{aicc_2025}; the Apple and
Stanford extractor study reports differences of up to ten percentage points
on WikiTQ from extractor choice alone \citep{beyond_single_extractor_2026};
lightweight extraction models continue to push in the same direction
\citep{dripper_2026}. We concede all of it. The same abstract that reports
the ten points also observes that different extractors may lead to similar
model performance on standard language-understanding tasks while the pages
surviving a fixed filtering pipeline differ substantially, which is this
paper's unpriced-loss claim stated by the field's own leading extraction
study, and stated in a refereed venue rather than a preprint. AICC's venue is
unverified on the arXiv record and we cite it as a preprint. One condition is
missing from that whole line, and it is the one this paper's mechanism turns
on: AICC's gain is measured against trafilatura-extracted text and never
against the same corpus under randomized notation, which is exactly the
format-augmentation condition. Whether markup fidelity or markup reliability
is doing the work in that result is, as far as we can tell, untested.

\textbf{Corpus measurement.} This paper is in the tradition of what is in a
corpus rather than what it scores \citep{wimbd_2024,dodge_2021,aboutme_2024,
quality_glance_2022,pile_datasheet_2022}. The nearest sibling profiles three
million Dolma passages on discourse dimensions with a trained scorer
\citep{narradolma_2026}; the boundary is that a trained scorer answers
questions a regular expression cannot, and a regular expression can be asked
of a laboratory. Fingerprinting work established that dataset-specific
formatting propagates through training into outputs
\citep{mansour_2025_fingerprints}, and register-based stratification is the
closest existing non-domain coordinate \citep{register_matters_2025}. Quality
filters have their own version of the problem: restyling a document into
Wikipedia's conventions, with facts and length held fixed, reverses an
educational-quality classifier on roughly seven percent of documents
\citep{klimaszewski_2026}.

\textbf{Mixing and augmentation.} The data-mixing literature optimizes
sampling weights over content-defined partitions
\citep{data_mixing_survey_2026,doremi_2023,doge_2024,regmix_2025,
mixing_laws_2025}, and every method in it operates over documents as built;
newer coordinates such as web-graph position are still properties of the
content or its provenance \citep{hubs_fringes_2026}. Augmentation research
rephrases at pre-training scale \citep{rephrasing_web_2024,beyondweb_2025,
synth_pretrain_2026} or diversifies document formats to make knowledge
extractable \citep{allenzhu_2023_physics31,zhu_2025_generalization}, and a
2026 taxonomy of pre-training augmentation has no notation axis in it at all
\citep{augmentation_taxonomy_2026}. Partial coverage does appear to suffice
in the analogous supervised setting, where expanding roughly 30\% of a
fine-tuning set into multiple answer formats recovers most of the gain of
full expansion \citep{multiformat_2026}; whether the same saturation holds
for notation at pre-training scale is untested, and it is one of the cheapest
open questions this program raises.

\textbf{Adjacent directions.} Rendering documents as pixels rather than
extracting them attacks the same loss from the other side, reporting that
visual pre-training on ``the same underlying corpora'' outperforms text-only
pre-training \citep{visual_pretraining_2026}, with a retrieval-time analogue
\citep{pixelrag_2026}; the ground is not empty. Encoder-era probing finds
long-document transformers acquire implicit representations of document
structure during pre-training and benefit from infusing more of it
\citep{buchmann_2024_docstructure}; that line adds structure to models and
measures neither the corpus nor the cost of the announcement.
Document-level organization
matters at mid-training: on the content-matched arm, holding generated text
and tokens fixed, book-level packaging is worth $+1.02$ on LLM-synthesized
textbook data, and a length-matched control rules out document length alone
\citep{tao_2026_booklevel}. Whitespace as load-bearing notation that
extraction destroys has been argued from the poetry side
\citep{whitespace_poets_2025}. Format effects at inference time are a
separate literature that we deliberately do not lean on
\citep{format_tax_2026,structural_attention_tax_2026,price_of_format_2025,
zhang_2025_formatbias}.

\section{Coda}
\label{sec:coda}

Nobody decided that the pre-training stream should stop demanding structural
inference. It fell out of a line filter here and a default output mode there,
a books subset quietly dropped, a long-context slot filled with whatever was
longest. Each step was locally correct and improved the number being watched,
and the cost landed on a dimension nobody had instrumented.

So: record it, then build it. Recording it costs a field on a card and a pass
over tokens that have already been counted once. Building it costs a
public-domain corpus, a deletion, a sidecar and a validator, aimed at a
training stage that is under one percent of the budget.

What we would ask a reader to take away is not the census or the null but the
choice they make visible. Every operator in the current pipeline is chosen
for how faithfully it preserves a document. At least one of them should be
chosen for what it teaches the model to do.

\bibliographystyle{plainnat}
\bibliography{references}

\begin{thebibliography}{94}
\providecommand{\natexlab}[1]{#1}
\providecommand{\url}[1]{\texttt{#1}}
\expandafter\ifx\csname urlstyle\endcsname\relax
  \providecommand{\doi}[1]{doi: #1}\else
  \providecommand{\doi}{doi: \begingroup \urlstyle{rm}\Url}\fi

\bibitem[{Allen Institute for AI}(2026)]{olmo_core_2026}
{Allen Institute for AI}.
\newblock {OLMo-core}: {\texttt{src/scripts/official/OLMo3/}}.
\newblock \url{https://github.com/allenai/OLMo-core}, 2026.
\newblock Read at commit \texttt{064b172} (2026-07-29).

\bibitem[Allen-Zhu and Li(2023)]{allenzhu_2023_physics31}
Zeyuan Allen-Zhu and Yuanzhi Li.
\newblock Physics of language models: Part 3.1, knowledge storage and
  extraction.
\newblock \emph{arXiv preprint arXiv:2309.14316}, 2023.

\bibitem[Altmann et~al.(2012)Altmann, Cristadoro, and
  Degli~Esposti]{altmann_2012}
Eduardo~G. Altmann, Giampaolo Cristadoro, and Mirko Degli~Esposti.
\newblock On the origin of long-range correlations in texts.
\newblock \emph{Proceedings of the National Academy of Sciences}, 109\penalty0
  (29):\penalty0 11582--11587, 2012.
\newblock DOI 10.1073/pnas.1117723109.

\bibitem[{\'A}lvarez-Lacalle et~al.(2006){\'A}lvarez-Lacalle, Dorow, Eckmann,
  and Moses]{alvarez_lacalle_2006}
E.~{\'A}lvarez-Lacalle, B.~Dorow, J.-P. Eckmann, and E.~Moses.
\newblock Hierarchical structures induce long-range dynamical correlations in
  written texts.
\newblock \emph{Proceedings of the National Academy of Sciences}, 103\penalty0
  (21):\penalty0 7956--7961, 2006.
\newblock DOI 10.1073/pnas.0510673103.

\bibitem[Aoyama et~al.(2026)Aoyama, Wilcox, and Schneider]{aoyama_2026}
Tatsuya Aoyama, Ethan~Gotlieb Wilcox, and Nathan Schneider.
\newblock Predicting the emergence of induction heads in language model
  pretraining.
\newblock \emph{arXiv preprint arXiv:2511.16893v3}, 2026.
\newblock Version-pinned: v3 (ICML 2026 camera-ready), Eq.~7.

\bibitem[Auer et~al.(2024)Auer, Lysak, Nassar, Dolfi, Livathinos, Vagenas,
  Berrospi~Ramis, Omenetti, Lindlbauer, Dinkla, et~al.]{docling_2024}
Christoph Auer, Maksym Lysak, Ahmed Nassar, Michele Dolfi, Nikolaos Livathinos,
  Panos Vagenas, Cesar Berrospi~Ramis, Matteo Omenetti, Fabian Lindlbauer,
  Kasper Dinkla, et~al.
\newblock Docling technical report.
\newblock \emph{arXiv preprint arXiv:2408.09869}, 2024.

\bibitem[Badoni et~al.(2026)Badoni, Chen, and Wang]{hubs_fringes_2026}
Vedant Badoni, Danqi Chen, and Xinyi Wang.
\newblock Hubs or fringes: Pretraining data selection via web graph centrality.
\newblock \emph{arXiv preprint arXiv:2606.11499}, 2026.

\bibitem[Baghaei~Potraghloo et~al.(2026)Baghaei~Potraghloo, Azizi, Kundu, and
  Pedram]{one_token_away_2026}
Erfan Baghaei~Potraghloo, Seyedarmin Azizi, Souvik Kundu, and Massoud Pedram.
\newblock One token away from collapse: The fragility of instruction-tuned
  helpfulness.
\newblock \emph{arXiv preprint arXiv:2604.13006}, 2026.

\bibitem[Barbaresi(2026)]{trafilatura_2026}
Adrien Barbaresi.
\newblock trafilatura: {\texttt{trafilatura/xml.py}}.
\newblock \url{https://github.com/adbar/trafilatura}, 2026.
\newblock Version 2.2.0; read at commit \texttt{c1bc953} (2026-07-31).

\bibitem[Bhyravajjula et~al.(2025)Bhyravajjula, Walsh, Preus, and
  Antoniak]{whitespace_poets_2025}
Sriharsh Bhyravajjula, Melanie Walsh, Anna Preus, and Maria Antoniak.
\newblock so much depends / upon / a whitespace: Why whitespace matters for
  poets and {LLM}s.
\newblock In \emph{Proceedings of the 2025 Conference on Empirical Methods in
  Natural Language Processing (EMNLP)}. Association for Computational
  Linguistics, 2025.
\newblock URL \url{https://aclanthology.org/2025.emnlp-main.1783/}.

\bibitem[Biderman et~al.(2022)Biderman, Bicheno, and Gao]{pile_datasheet_2022}
Stella Biderman, Kieran Bicheno, and Leo Gao.
\newblock Datasheet for the {Pile}.
\newblock \emph{arXiv preprint arXiv:2201.07311}, 2022.

\bibitem[Buchmann et~al.(2024)Buchmann, Eichler, Bodensohn, Kuznetsov, and
  Gurevych]{buchmann_2024_docstructure}
Jan Buchmann, Max Eichler, Jan-Micha Bodensohn, Ilia Kuznetsov, and Iryna
  Gurevych.
\newblock Document structure in long document transformers.
\newblock In \emph{Proceedings of the 18th Conference of the European Chapter
  of the Association for Computational Linguistics (EACL), Volume 1: Long
  Papers}, pages 1056--1073, 2024.
\newblock URL \url{https://aclanthology.org/2024.eacl-long.64/}.

\bibitem[Cargnelutti et~al.(2025)Cargnelutti, Brobston, Hess, Cushman, Mukk,
  Scourtas, Courtney, Leppert, Watson, Whitehead, and
  Zittrain]{institutional_books_2025}
Matteo Cargnelutti, Catherine Brobston, John Hess, Jack Cushman, Kristi Mukk,
  Aristana Scourtas, Kyle Courtney, Greg Leppert, Amanda Watson, Martha
  Whitehead, and Jonathan Zittrain.
\newblock Institutional books 1.0: A 242{B} token dataset from {Harvard
  Library}'s collections, refined for accuracy and usability.
\newblock \emph{arXiv preprint arXiv:2506.08300}, 2025.

\bibitem[Carlsson et~al.(2025)Carlsson, Liu, Ward, Kurfali, and
  Nivre]{carlsson_2025_hyperfitting}
Fredrik Carlsson, Fangyu Liu, Daniel Ward, Murathan Kurfali, and Joakim Nivre.
\newblock The hyperfitting phenomenon: Sharpening and stabilizing {LLM}s for
  open-ended text generation.
\newblock In \emph{International Conference on Learning Representations
  (ICLR)}, 2025.

\bibitem[Chen et~al.(2025)Chen, Wu, Xu, and Zhang]{ladm_2025}
Jianghao Chen, Junhong Wu, Yangyifan Xu, and Jiajun Zhang.
\newblock {LADM}: Long-context training data selection with attention-based
  dependency measurement for {LLM}s.
\newblock In \emph{Proceedings of the 63rd Annual Meeting of the Association
  for Computational Linguistics (ACL)}, 2025.

\bibitem[Chen et~al.(2026{\natexlab{a}})Chen, Zhang, Bai, Hu, and
  Wang]{augmentation_taxonomy_2026}
Michael~K. Chen, Xikun Zhang, Fan Bai, Zhengding Hu, and Zhen Wang.
\newblock Demystifying training-time augmentation for data-constrained language
  model pretraining.
\newblock \emph{arXiv preprint arXiv:2606.16246}, 2026{\natexlab{a}}.

\bibitem[Chen et~al.(2026{\natexlab{b}})Chen, Miao, Supryadi, and
  Xiong]{data_mixing_survey_2026}
Zhuo Chen, Yuxuan Miao, Supryadi, and Deyi Xiong.
\newblock Data mixing for large language models pretraining: A survey and
  outlook.
\newblock \emph{Data Intelligence}, 8, 2026{\natexlab{b}}.

\bibitem[{DatologyAI}(2025)]{beyondweb_2025}
{DatologyAI}.
\newblock {BeyondWeb}: Lessons from scaling synthetic data for trillion-scale
  pretraining.
\newblock \emph{arXiv preprint arXiv:2508.10975}, 2025.

\bibitem[{DeepSeek-AI}(2024)]{deepseekv3_2024}
{DeepSeek-AI}.
\newblock {DeepSeek-V3} technical report.
\newblock \emph{arXiv preprint arXiv:2412.19437}, 2024.

\bibitem[Deng et~al.(2026)Deng, Lin, Lin, Liu, Sun, Ma, and
  Gong]{longrange_info_2026}
Haoran Deng, Yingyu Lin, Zhenghao Lin, Xiao Liu, Yizhou Sun, Yi-An Ma, and
  Yeyun Gong.
\newblock Beyond length: Quantifying long-range information for long-context
  {LLM} pretraining data.
\newblock In \emph{International Conference on Learning Representations
  (ICLR)}, 2026.
\newblock Venue confirmed via OpenReview; not on the arXiv record.

\bibitem[Dodge et~al.(2021)Dodge, Sap, Maraso{v}i{\'c}, Agnew, Ilharco,
  Groeneveld, Mitchell, and Gardner]{dodge_2021}
Jesse Dodge, Maarten Sap, Ana Maraso{v}i{\'c}, William Agnew, Gabriel Ilharco,
  Dirk Groeneveld, Margaret Mitchell, and Matt Gardner.
\newblock Documenting large webtext corpora: A case study on the {Colossal
  Clean Crawled Corpus}.
\newblock In \emph{Proceedings of the 2021 Conference on Empirical Methods in
  Natural Language Processing (EMNLP)}, 2021.

\bibitem[Elazar et~al.(2024)Elazar, Bhagia, Magnusson, Ravichander, Schwenk,
  Suhr, Walsh, Groeneveld, Soldaini, Singh, Hajishirzi, Smith, and
  Dodge]{wimbd_2024}
Yanai Elazar, Akshita Bhagia, Ian Magnusson, Abhilasha Ravichander, Dustin
  Schwenk, Alane Suhr, Pete Walsh, Dirk Groeneveld, Luca Soldaini, Sameer
  Singh, Hanna Hajishirzi, Noah~A. Smith, and Jesse Dodge.
\newblock What's in my big data?
\newblock In \emph{International Conference on Learning Representations
  (ICLR)}, 2024.
\newblock Spotlight.

\bibitem[Fan et~al.(2024)Fan, Pagliardini, and Jaggi]{doge_2024}
Simin Fan, Matteo Pagliardini, and Martin Jaggi.
\newblock {DoGE}: Domain reweighting with generalization estimation.
\newblock \emph{arXiv preprint arXiv:2310.15393}, 2024.

\bibitem[Freeburg(2026)]{freeburg_2026}
E.~M. Freeburg.
\newblock The last fingerprint: How {Markdown} training shapes {LLM} prose.
\newblock \emph{arXiv preprint arXiv:2603.27006}, 2026.

\bibitem[Gao et~al.(2025)Gao, Wu, Lin, Zhang, and Hu]{nextlong_2025}
Chaochen Gao, Xing Wu, Zijia Lin, Debing Zhang, and Songlin Hu.
\newblock {NExtLong}: Toward effective long-context training without long
  documents.
\newblock In \emph{Proceedings of the 42nd International Conference on Machine
  Learning (ICML)}, 2025.

\bibitem[Gao et~al.(2020)Gao, Biderman, Black, Golding, Hoppe, Foster, Phang,
  He, Thite, Nabeshima, Presser, and Leahy]{pile_2020}
Leo Gao, Stella Biderman, Sid Black, Laurence Golding, Travis Hoppe, Charles
  Foster, Jason Phang, Horace He, Anish Thite, Noa Nabeshima, Shawn Presser,
  and Connor Leahy.
\newblock The {Pile}: An 800{GB} dataset of diverse text for language modeling.
\newblock \emph{arXiv preprint arXiv:2101.00027}, 2020.

\bibitem[Geirhos et~al.(2020)Geirhos, Jacobsen, Michaelis, Zemel, Brendel,
  Bethge, and Wichmann]{geirhos_2020}
Robert Geirhos, J{\"o}rn-Henrik Jacobsen, Claudio Michaelis, Richard Zemel,
  Wieland Brendel, Matthias Bethge, and Felix~A. Wichmann.
\newblock Shortcut learning in deep neural networks.
\newblock \emph{Nature Machine Intelligence}, 2:\penalty0 665--673, 2020.
\newblock DOI 10.1038/s42256-020-00257-z.

\bibitem[Hermann et~al.(2024)Hermann, Mobahi, Fel, and Mozer]{hermann_2024}
Katherine~L. Hermann, Hossein Mobahi, Thomas Fel, and Michael~C. Mozer.
\newblock On the foundations of shortcut learning.
\newblock In \emph{International Conference on Learning Representations
  (ICLR)}, 2024.

\bibitem[Holtzman et~al.(2020)Holtzman, Buys, Du, Forbes, and
  Choi]{holtzman_2020_degeneration}
Ari Holtzman, Jan Buys, Li~Du, Maxwell Forbes, and Yejin Choi.
\newblock The curious case of neural text degeneration.
\newblock In \emph{International Conference on Learning Representations
  (ICLR)}, 2020.

\bibitem[{Hugging Face}(2025{\natexlab{a}})]{finepdfs_2025}
{Hugging Face}.
\newblock {FinePDFs} dataset card.
\newblock \url{https://huggingface.co/datasets/HuggingFaceFW/finepdfs},
  2025{\natexlab{a}}.
\newblock Accessed 2026-08-09.

\bibitem[{Hugging Face}(2025{\natexlab{b}})]{smollm3_2025}
{Hugging Face}.
\newblock {SmolLM3}: smol, multilingual, long-context reasoner.
\newblock \url{https://huggingface.co/blog/smollm3}, 2025{\natexlab{b}}.
\newblock Accessed 2026-08-09.

\bibitem[{Hugging Face}(2026)]{datatrove_2026}
{Hugging Face}.
\newblock datatrove:
  {\texttt{src/datatrove/pipeline/extractors/trafilatura.py}}.
\newblock \url{https://github.com/huggingface/datatrove}, 2026.
\newblock Read at commit \texttt{0eb8e30} (2026-08-06).

\bibitem[Idahl et~al.(2026)Idahl, Droste, Pl{\"u}ster, and
  Harries]{propella_2026}
Maximilian Idahl, Benedikt Droste, Bj{\"o}rn Pl{\"u}ster, and Jan~Philipp
  Harries.
\newblock propella-1: Multi-property document annotation for {LLM} data
  curation at scale.
\newblock \emph{arXiv preprint arXiv:2602.12414}, 2026.
\newblock Preprint; no venue on the arXiv record.

\bibitem[{International Organization for Standardization}(2008)]{iso32000}
{International Organization for Standardization}.
\newblock {ISO} 32000-1:2008 --- document management: Portable document format
  --- part 1: {PDF} 1.7.
\newblock \url{https://www.iso.org/standard/51502.html}, 2008.

\bibitem[Jia et~al.(2025)Jia, Chen, Wu, Gao, Lin, Zhang, Hu, and
  Guo]{entropylong_2025}
Junlong Jia, Ziyang Chen, Xing Wu, Chaochen Gao, Zijia Lin, Debing Zhang,
  Songlin Hu, and Binghui Guo.
\newblock {EntropyLong}: Effective long-context training via predictive
  uncertainty.
\newblock \emph{arXiv preprint arXiv:2510.02330}, 2025.

\bibitem[Johnson et~al.(2026)Johnson, Ash, Piper, and
  Antoniak]{narradolma_2026}
Teagan Johnson, Elliott Ash, Andrew Piper, and Maria Antoniak.
\newblock Characterizing narrative content in web-scale {LLM} pretraining data.
\newblock \emph{arXiv preprint arXiv:2606.19468}, 2026.

\bibitem[Kandpal et~al.(2025)Kandpal, Lester, Raffel, Majstorovic, Biderman,
  Abbasi, Soldaini, Shippole, Cooper, Skowron, et~al.]{common_pile_2025}
Nikhil Kandpal, Brian Lester, Colin Raffel, Sebastian Majstorovic, Stella
  Biderman, Baber Abbasi, Luca Soldaini, Enrico Shippole, A.~Feder Cooper,
  Aviya Skowron, et~al.
\newblock The {Common Pile} v0.1: An 8{TB} dataset of public domain and openly
  licensed text.
\newblock \emph{arXiv preprint arXiv:2506.05209}, 2025.

\bibitem[Karimi~Mahabadi et~al.(2025)Karimi~Mahabadi, Satheesh, Prabhumoye,
  Patwary, Shoeybi, and Catanzaro]{nemotroncc_math_2025}
Rabeeh Karimi~Mahabadi, Sanjeev Satheesh, Shrimai Prabhumoye, Mostofa Patwary,
  Mohammad Shoeybi, and Bryan Catanzaro.
\newblock {Nemotron-CC-Math}: A 133 billion-token-scale high quality math
  pretraining dataset.
\newblock \emph{arXiv preprint arXiv:2508.15096}, 2025.

\bibitem[Klimaszewski and Andruszkiewicz(2026)]{klimaszewski_2026}
Mateusz Klimaszewski and Piotr Andruszkiewicz.
\newblock Is a document educational or just {Wikipedia}-style? pitfalls of
  classifier-based quality filtering.
\newblock In \emph{Proceedings of the 64th Annual Meeting of the Association
  for Computational Linguistics (ACL)}, 2026.

\bibitem[Kreutzer et~al.(2022)Kreutzer, Caswell, Wang, Wahab, van Esch,
  Ulzii-Orshikh, Tapo, Subramani, Sokolov, Sikasote,
  et~al.]{quality_glance_2022}
Julia Kreutzer, Isaac Caswell, Lisa Wang, Ahsan Wahab, Daan van Esch,
  Nasanbayar Ulzii-Orshikh, Allahsera Tapo, Nishant Subramani, Artem Sokolov,
  Claytone Sikasote, et~al.
\newblock Quality at a glance: An audit of web-crawled multilingual datasets.
\newblock \emph{Transactions of the Association for Computational Linguistics},
  10:\penalty0 50--72, 2022.

\bibitem[Lee et~al.(2026)Lee, D'Antoni, and Berg-Kirkpatrick]{format_tax_2026}
Ivan~Yee Lee, Loris D'Antoni, and Taylor Berg-Kirkpatrick.
\newblock The format tax.
\newblock \emph{arXiv preprint arXiv:2604.03616}, 2026.

\bibitem[Li et~al.(2024)Li, Fang, Smyrnis, Ivgi, Jordan, Gadre, Bansal, Guha,
  Keh, Arora, et~al.]{dclm_2024}
Jeffrey Li, Alex Fang, Georgios Smyrnis, Maor Ivgi, Matt Jordan, Samir Gadre,
  Hritik Bansal, Etash Guha, Sedrick Keh, Kushal Arora, et~al.
\newblock {DataComp-LM}: In search of the next generation of training sets for
  language models.
\newblock In \emph{Advances in Neural Information Processing Systems (NeurIPS),
  Datasets and Benchmarks Track}, 2024.

\bibitem[Li et~al.(2026{\natexlab{a}})Li, Gardner, Kang, Shi, Singh, Li,
  Shandilya, Hall, Tuzel, Liang, Schmidt, Pouransari, and
  Faghri]{beyond_single_extractor_2026}
Jeffrey Li, Joshua~P. Gardner, Doug Kang, Fangping Shi, Karanjeet Singh,
  Chun-Liang Li, Herumb Shandilya, David Leo~Wright Hall, Oncel Tuzel, Percy
  Liang, Ludwig Schmidt, Hadi Pouransari, and Fartash Faghri.
\newblock Beyond a single extractor: Re-thinking {HTML}-to-text extraction for
  {LLM} pre-training.
\newblock In \emph{Findings of the Association for Computational Linguistics:
  EACL 2026}. Association for Computational Linguistics, 2026{\natexlab{a}}.
\newblock URL \url{https://aclanthology.org/2026.findings-eacl.307/}.

\bibitem[Li et~al.(2026{\natexlab{b}})Li, Zhu, Wu, Bao, and
  Evans]{narrative_flattening_2026}
Zehan Li, Yutong Zhu, Siyang Wu, Honglin Bao, and James~A. Evans.
\newblock Narrative flattening: How post-training compresses thematic,
  affective, and stylistic variation in {LLM} fiction.
\newblock \emph{arXiv preprint arXiv:2605.27878}, 2026{\natexlab{b}}.

\bibitem[Lin and Tegmark(2017)]{lin_tegmark_2017}
Henry~W. Lin and Max Tegmark.
\newblock Criticality in formal languages and statistical physics.
\newblock \emph{Entropy}, 19\penalty0 (7):\penalty0 299, 2017.

\bibitem[Liu et~al.(2026{\natexlab{a}})Liu, Zheng, Cao, Jin, Cui, and
  Zhou]{multiformat_2026}
June~M. Liu, Shaomian Zheng, He~Cao, Dingnan Jin, Qing Cui, and Jun Zhou.
\newblock Improving cross-format robustness in language models with
  multi-format training.
\newblock \emph{arXiv preprint arXiv:2606.11643}, 2026{\natexlab{a}}.

\bibitem[Liu et~al.(2026{\natexlab{b}})Liu, Peng, Ning, Chu, Qiu, Ma, Zhu, Min,
  Lu, Hou, et~al.]{dripper_2026}
Mengjie Liu, Jiahui Peng, Wenchang Ning, Pei Chu, Jiantao Qiu, Ren Ma, He~Zhu,
  Rui Min, Lindong Lu, Linfeng Hou, et~al.
\newblock {Dripper}: Token-efficient main {HTML} extraction with a lightweight
  {LM}.
\newblock \emph{arXiv preprint arXiv:2511.23119}, 2026{\natexlab{b}}.

\bibitem[Liu et~al.(2025)Liu, Zheng, Muennighoff, Zeng, Dou, Pang, Jiang, and
  Lin]{regmix_2025}
Qian Liu, Xiaosen Zheng, Niklas Muennighoff, Guangtao Zeng, Longxu Dou, Tianyu
  Pang, Jing Jiang, and Min Lin.
\newblock {RegMix}: Data mixture as regression for language model pre-training.
\newblock In \emph{International Conference on Learning Representations
  (ICLR)}, 2025.

\bibitem[Liu and Chu(2026)]{entrainment_2026}
Yang Liu and Chenhui Chu.
\newblock Sentence-level contextual entrainment in large language models.
\newblock \emph{arXiv preprint arXiv:2606.24077}, 2026.

\bibitem[{Llama Team, AI @ Meta}(2024)]{llama3_2024}
{Llama Team, AI @ Meta}.
\newblock The {Llama 3} herd of models.
\newblock \emph{arXiv preprint arXiv:2407.21783}, 2024.

\bibitem[Lucy et~al.(2024)Lucy, Gururangan, Soldaini, Strubell, Bamman, Klein,
  and Dodge]{aboutme_2024}
Li~Lucy, Suchin Gururangan, Luca Soldaini, Emma Strubell, David Bamman,
  Lauren~F. Klein, and Jesse Dodge.
\newblock {AboutMe}: Using self-descriptions in webpages to document the
  effects of {English} pretraining data filters.
\newblock In \emph{Proceedings of the 62nd Annual Meeting of the Association
  for Computational Linguistics (ACL)}, 2024.

\bibitem[Ma et~al.(2025)Ma, Qiu, Xu, Chu, Liu, Ren, Qu, Peng, Hou, Liu,
  et~al.]{aicc_2025}
Ren Ma, Jiantao Qiu, Chao Xu, Pei Chu, Kaiwen Liu, Pengli Ren, Yuan Qu, Jiahui
  Peng, Linfeng Hou, Mengjie Liu, et~al.
\newblock {AICC}: Parse {HTML} finer, make models better --- a 7.3{T}
  {AI}-ready corpus built by a model-based {HTML} parser.
\newblock \emph{arXiv preprint arXiv:2511.16397}, 2025.

\bibitem[Maini et~al.(2024)Maini, Seto, Bai, Grangier, Zhang, and
  Jaitly]{rephrasing_web_2024}
Pratyush Maini, Skyler Seto, He~Bai, David Grangier, Yizhe Zhang, and Navdeep
  Jaitly.
\newblock Rephrasing the web: A recipe for compute and data-efficient language
  modeling.
\newblock \emph{arXiv preprint arXiv:2401.16380}, 2024.

\bibitem[Mansour and Heckel(2025)]{mansour_2025_fingerprints}
Youssef Mansour and Reinhard Heckel.
\newblock Measuring fingerprints of web-filtered text datasets and fingerprint
  propagation through training.
\newblock In \emph{Advances in Neural Information Processing Systems
  (NeurIPS)}, 2025.
\newblock Spotlight.

\bibitem[Mikhaylovskiy and Churilov(2023)]{mikhaylovskiy_2023}
Nikolay Mikhaylovskiy and Ilya Churilov.
\newblock Autocorrelations decay in texts and applicability limits of language
  models.
\newblock In \emph{Computational Linguistics and Intellectual Technologies
  (Dialogue)}, 2023.

\bibitem[Myntti et~al.(2025)Myntti, Henriksson, Laippala, and
  Pyysalo]{register_matters_2025}
Amanda Myntti, Erik Henriksson, Veronika Laippala, and Sampo Pyysalo.
\newblock Register always matters: Analysis of {LLM} pretraining data through
  the lens of language variation.
\newblock In \emph{Proceedings of the Second Conference on Language Modeling
  (COLM)}, 2025.

\bibitem[Niklaus et~al.(2026)Niklaus, Yamaguchi, {\v{S}}tef{\'a}nik, Penedo,
  Kydl{\'\i}{\v{c}}ek, Bakouch, Tunstall, Beeching, Frere, Raffel, von Werra,
  and Wolf]{synth_pretrain_2026}
Joel Niklaus, Atsuki Yamaguchi, Michal {\v{S}}tef{\'a}nik, Guilherme Penedo,
  Hynek Kydl{\'\i}{\v{c}}ek, Elie Bakouch, Lewis Tunstall, Edward~Emanuel
  Beeching, Thibaud Frere, Colin Raffel, Leandro von Werra, and Thomas Wolf.
\newblock How can we synthesize high-quality pretraining data? a systematic
  study of prompt design, generator model, and source data.
\newblock In \emph{Conference on Language Modeling (COLM)}, 2026.

\bibitem[Paul et~al.(2026)Paul, Helm, Glava{\v{s}}, and
  Gurevych]{octolong_2026}
Indraneil Paul, Falko Helm, Goran Glava{\v{s}}, and Iryna Gurevych.
\newblock {OctoLong}: Mid-training on cross-repository code contexts enhances
  long-context modeling.
\newblock \emph{arXiv preprint arXiv:2608.05141}, 2026.

\bibitem[Penedo et~al.(2024)Penedo, Kydl{\'\i}{\v{c}}ek, Ben~Allal, Lozhkov,
  Mitchell, Raffel, Von~Werra, and Wolf]{fineweb_2024}
Guilherme Penedo, Hynek Kydl{\'\i}{\v{c}}ek, Loubna Ben~Allal, Anton Lozhkov,
  Margaret Mitchell, Colin Raffel, Leandro Von~Werra, and Thomas Wolf.
\newblock The {FineWeb} datasets: Decanting the web for the finest text data at
  scale.
\newblock In \emph{Advances in Neural Information Processing Systems (NeurIPS),
  Datasets and Benchmarks Track}, 2024.

\bibitem[Pethe et~al.(2020)Pethe, Kim, and Skiena]{chaptercaptor_2020}
Charuta Pethe, Allen Kim, and Steven Skiena.
\newblock Chapter captor: Text segmentation in novels.
\newblock In \emph{Proceedings of the 2020 Conference on Empirical Methods in
  Natural Language Processing (EMNLP)}, pages 8373--8383. Association for
  Computational Linguistics, 2020.
\newblock URL \url{https://aclanthology.org/2020.emnlp-main.672/}.

\bibitem[Poznanski et~al.(2025)Poznanski, Rangapur, Borchardt, Dunkelberger,
  Huff, Lin, Wilhelm, Lo, and Soldaini]{olmocr_2025}
Jake Poznanski, Aman Rangapur, Jon Borchardt, Jason Dunkelberger, Regan Huff,
  Daniel Lin, Christopher Wilhelm, Kyle Lo, and Luca Soldaini.
\newblock {olmOCR}: Unlocking trillions of tokens in {PDF}s with vision
  language models.
\newblock \emph{arXiv preprint arXiv:2502.18443}, 2025.

\bibitem[Presser(2020)]{presser_epub2txt}
Shawn Presser.
\newblock \texttt{epub2txt-all} (the {Books3} converter).
\newblock \url{https://github.com/shawwn/scrap}, 2020.
\newblock Read at commit \texttt{31fd266} (2026-07-30).

\bibitem[{Qwen Team}(2025)]{qwen3_2025}
{Qwen Team}.
\newblock {Qwen3} technical report.
\newblock \emph{arXiv preprint arXiv:2505.09388}, 2025.

\bibitem[Rae et~al.(2019)Rae, Potapenko, Jayakumar, and Lillicrap]{pg19_2019}
Jack~W. Rae, Anna Potapenko, Siddhant~M. Jayakumar, and Timothy~P. Lillicrap.
\newblock Compressive transformers for long-range sequence modelling.
\newblock \emph{arXiv preprint arXiv:1911.05507}, 2019.

\bibitem[Raffel et~al.(2020)Raffel, Shazeer, Roberts, Lee, Narang, Matena,
  Zhou, Li, and Liu]{t5_c4_2020}
Colin Raffel, Noam Shazeer, Adam Roberts, Katherine Lee, Sharan Narang, Michael
  Matena, Yanqi Zhou, Wei Li, and Peter~J. Liu.
\newblock Exploring the limits of transfer learning with a unified text-to-text
  transformer.
\newblock \emph{Journal of Machine Learning Research}, 21\penalty0
  (140):\penalty0 1--67, 2020.

\bibitem[{Reducto AI}(2025)]{rolmocr_2025}
{Reducto AI}.
\newblock {RolmOCR} model card.
\newblock \url{https://huggingface.co/reducto/RolmOCR}, 2025.
\newblock Accessed 2026-08-09.

\bibitem[Reynolds-Cu{\'e}llar et~al.(2026)Reynolds-Cu{\'e}llar, Wong-Villacres,
  Alvarado~Garcia, and Precel]{doc_tools_review_2026}
Pedro Reynolds-Cu{\'e}llar, Marisol Wong-Villacres, Adriana Alvarado~Garcia,
  and Heila Precel.
\newblock From reflection to repair: A scoping review of dataset documentation
  tools.
\newblock \emph{arXiv preprint arXiv:2602.15968}, 2026.

\bibitem[Sarkar and Howard(2019)]{sarkar_howard_2019}
Aakash Sarkar and Marc Howard.
\newblock Scale-dependent relationships in natural language.
\newblock \emph{arXiv preprint arXiv:1912.07506}, 2019.

\bibitem[Shen(2019)]{shen_2019_mi}
Huitao Shen.
\newblock Mutual information scaling and expressive power of sequence models.
\newblock \emph{arXiv preprint arXiv:1905.04271}, 2019.

\bibitem[Soldaini et~al.(2024)Soldaini, Kinney, Bhagia, Schwenk, Atkinson,
  Authur, Bogin, Chandu, Dumas, Elazar, et~al.]{dolma_2024}
Luca Soldaini, Rodney Kinney, Akshita Bhagia, Dustin Schwenk, David Atkinson,
  Russell Authur, Ben Bogin, Khyathi Chandu, Jennifer Dumas, Yanai Elazar,
  et~al.
\newblock {Dolma}: an open corpus of three trillion tokens for language model
  pretraining research.
\newblock In \emph{Proceedings of the 62nd Annual Meeting of the Association
  for Computational Linguistics (ACL)}, 2024.

\bibitem[Su et~al.(2025)Su, Kong, Lin, Jennings, Norick, Kliegl, Patwary,
  Shoeybi, and Catanzaro]{nemotroncc_2025}
Dan Su, Kezhi Kong, Ying Lin, Joseph Jennings, Brandon Norick, Markus Kliegl,
  Mostofa Patwary, Mohammad Shoeybi, and Bryan Catanzaro.
\newblock {Nemotron-CC}: Transforming {Common Crawl} into a refined
  long-horizon pretraining dataset.
\newblock In \emph{Proceedings of the 63rd Annual Meeting of the Association
  for Computational Linguistics (ACL)}, 2025.

\bibitem[Sun et~al.(2025)Sun, Yin, Xu, Kolter, and
  Liu]{sun_2025_idiosyncrasies}
Mingjie Sun, Yida Yin, Zhiqiu Xu, J.~Zico Kolter, and Zhuang Liu.
\newblock Idiosyncrasies in large language models.
\newblock In \emph{Proceedings of the 42nd International Conference on Machine
  Learning (ICML)}, 2025.

\bibitem[Sun et~al.(2022)Sun, Thai, and Iyyer]{chapterbreak_2022}
Simeng Sun, Katherine Thai, and Mohit Iyyer.
\newblock {ChapterBreak}: A challenge dataset for long-range language models.
\newblock In \emph{Proceedings of the 2022 Conference of the North American
  Chapter of the Association for Computational Linguistics: Human Language
  Technologies (NAACL-HLT)}, pages 3704--3714, Seattle, United States, 2022.
  Association for Computational Linguistics.
\newblock URL \url{https://aclanthology.org/2022.naacl-main.271/}.

\bibitem[Swartz et~al.(2024)Swartz, Balakrishnan, et~al.]{html2text}
Aaron Swartz, Chris Balakrishnan, et~al.
\newblock \texttt{html2text}: Convert {HTML} to {Markdown}-formatted text.
\newblock \url{https://github.com/Alir3z4/html2text}, 2024.
\newblock Accessed 2026-08-09.

\bibitem[Tao et~al.(2026)Tao, Peng, Li, Yuan, Wu, Yu, Wang, Chen, Yang, and
  Pan]{tao_2026_booklevel}
Jiawen Tao, Miao Peng, Yaoming Li, Xiaokun Yuan, Mengzhou Wu, Wenhan Yu, Guoan
  Wang, Nuo Chen, Tong Yang, and Maxm Pan.
\newblock Beyond rephrasing: Book-level organization improves synthetic
  textbook data for mid-training.
\newblock \emph{arXiv preprint arXiv:2607.28109}, 2026.

\bibitem[{Team Olmo}(2026)]{olmo3_2026}
{Team Olmo}.
\newblock {Olmo} 3.
\newblock \emph{arXiv preprint arXiv:2512.13961}, 2026.

\bibitem[Touvron et~al.(2023)Touvron, Lavril, Izacard, Martinet, Lachaux,
  Lacroix, Rozi{\`e}re, Goyal, Hambro, Azhar, et~al.]{llama1_2023}
Hugo Touvron, Thibaut Lavril, Gautier Izacard, Xavier Martinet, Marie-Anne
  Lachaux, Timoth{\'e}e Lacroix, Baptiste Rozi{\`e}re, Naman Goyal, Eric
  Hambro, Faisal Azhar, et~al.
\newblock {LLaMA}: Open and efficient foundation language models.
\newblock \emph{arXiv preprint arXiv:2302.13971}, 2023.

\bibitem[Vegeto(2026)]{right_reset_2026}
Mike Vegeto.
\newblock Right reset: Chunking by prefix removal.
\newblock \emph{arXiv preprint arXiv:2608.04330}, 2026.
\newblock Preprint.

\bibitem[Wang et~al.(2024)Wang, Xu, Zhao, Ouyang, Wu, Zhao, Xu, Liu, Qu, Shang,
  et~al.]{mineru_2024}
Bin Wang, Chao Xu, Xiaomeng Zhao, Linke Ouyang, Fan Wu, Zhiyuan Zhao, Rui Xu,
  Kaiwen Liu, Yuan Qu, Fukai Shang, et~al.
\newblock {MinerU}: An open-source solution for precise document content
  extraction.
\newblock \emph{arXiv preprint arXiv:2409.18839}, 2024.

\bibitem[Wang et~al.(2026)Wang, Li, Wang, Teiletche, Jin, Zaharia, Gonzalez,
  and Min]{pixelrag_2026}
Yichuan Wang, Zhifei Li, Zirui Wang, Paul Teiletche, Lesheng Jin, Matei
  Zaharia, Joseph~E. Gonzalez, and Sewon Min.
\newblock {PIXELRAG}: Web screenshots beat text for retrieval-augmented
  generation.
\newblock \emph{arXiv preprint arXiv:2606.28344}, 2026.

\bibitem[Winata et~al.(2025)Winata, Anugraha, Liu, Aji, Hung, Parashar, Irawan,
  Zhang, Yong, Cruz, Muennighoff, et~al.]{datarubrics_2025}
Genta~Indra Winata, David Anugraha, Emmy Liu, Alham~Fikri Aji, Shou-Yi Hung,
  Aditya Parashar, Patrick~Amadeus Irawan, Ruochen Zhang, Zheng-Xin Yong, Jan
  Christian~Blaise Cruz, Niklas Muennighoff, et~al.
\newblock Datasheets aren't enough: {DataRubrics} for automated quality metrics
  and accountability.
\newblock \emph{arXiv preprint arXiv:2506.01789}, 2025.

\bibitem[Wu et~al.(2025)Wu, Zhu, Zhao, Yu, Ran, Wong, Sun, and
  Li]{longattn_2025}
Longyun Wu, Dawei Zhu, Guangxiang Zhao, Zhuocheng Yu, Junfeng Ran, Xiangyu
  Wong, Lin Sun, and Sujian Li.
\newblock {LongAttn}: Selecting long-context training data via token-level
  attention.
\newblock \emph{arXiv preprint arXiv:2502.16860}, 2025.

\bibitem[Wu et~al.(2024)Wu, Morris, and Levine]{wu_2024_precaching}
Wilson Wu, John~X. Morris, and Lionel Levine.
\newblock Do language models plan ahead for future tokens?
\newblock In \emph{Conference on Language Modeling (COLM)}, 2024.

\bibitem[Xie et~al.(2023)Xie, Pham, Dong, Du, Liu, Lu, Liang, Le, Ma, and
  Yu]{doremi_2023}
Sang~Michael Xie, Hieu Pham, Xuanyi Dong, Nan Du, Hanxiao Liu, Yifeng Lu, Percy
  Liang, Quoc~V. Le, Tengyu Ma, and Adams~Wei Yu.
\newblock {DoReMi}: Optimizing data mixtures speeds up language model
  pretraining.
\newblock In \emph{Advances in Neural Information Processing Systems
  (NeurIPS)}, 2023.

\bibitem[Xu et~al.(2022)Xu, Liu, Yan, Cai, Li, and Li]{xu_2022_break_loop}
Jin Xu, Xiaojiang Liu, Jianhao Yan, Deng Cai, Huayang Li, and Jian Li.
\newblock Learning to break the loop: Analyzing and mitigating repetitions for
  neural text generation.
\newblock In \emph{Advances in Neural Information Processing Systems
  (NeurIPS)}, 2022.

\bibitem[Ye et~al.(2025)Ye, Liu, Sun, Zhan, Zhou, and Qiu]{mixing_laws_2025}
Jiasheng Ye, Peiju Liu, Tianxiang Sun, Jun Zhan, Yunhua Zhou, and Xipeng Qiu.
\newblock Data mixing laws: Optimizing data mixtures by predicting language
  modeling performance.
\newblock In \emph{International Conference on Learning Representations
  (ICLR)}, 2025.

\bibitem[Yun et~al.(2025)Yun, An, Wang, Peng, and Shang]{price_of_format_2025}
Longfei Yun, Chenyang An, Zilong Wang, Letian Peng, and Jingbo Shang.
\newblock The price of format: Diversity collapse in {LLM}s.
\newblock \emph{arXiv preprint arXiv:2505.18949}, 2025.

\bibitem[Zacks et~al.(2007)Zacks, Speer, Swallow, Braver, and
  Reynolds]{zacks_2007}
Jeffrey~M. Zacks, Nicole~K. Speer, Khena~M. Swallow, Todd~S. Braver, and
  Jeremy~R. Reynolds.
\newblock Event perception: A mind-brain perspective.
\newblock \emph{Psychological Bulletin}, 133\penalty0 (2):\penalty0 273--293,
  2007.
\newblock Cited as motivation, not as evidence, for the claim that discourse
  segmentation is constructed by the reader.

\bibitem[Zhang et~al.(2025)Zhang, Xiong, Chen, Zhou, Huang, and
  Zhang]{zhang_2025_formatbias}
Xuanchang Zhang, Wei Xiong, Lichang Chen, Tianyi Zhou, Heng Huang, and Tong
  Zhang.
\newblock From lists to emojis: How format bias affects model alignment.
\newblock In \emph{Proceedings of the 63rd Annual Meeting of the Association
  for Computational Linguistics (ACL)}, pages 26940--26961, 2025.

\bibitem[Zhang et~al.(2026)Zhang, Zhao, Zhang, Zhao, Lin, Tang, Zhou, Song,
  Liu, Ye, et~al.]{visual_pretraining_2026}
Yiming Zhang, Zhonghan Zhao, Wenwei Zhang, Haiteng Zhao, Tianyang Lin, Huanze
  Tang, Yunhua Zhou, Demin Song, Kuikun Liu, Haochen Ye, et~al.
\newblock Scalable visual pretraining for language intelligence.
\newblock \emph{arXiv preprint arXiv:2607.09657}, 2026.

\bibitem[Zhang and Zhang(2026)]{structural_attention_tax_2026}
Yuqi Zhang and Di~Zhang.
\newblock The structural attention tax: How retrieval format hijacks in-context
  learning independent of content.
\newblock \emph{arXiv preprint arXiv:2606.11198}, 2026.
\newblock Preprint.

\bibitem[Zhu et~al.(2026)Zhu, Li, Zou, Fu, Wang, He, and
  Yang]{effective_context_2026}
Jinchang Zhu, Jindong Li, Chengyu Zou, Rong Fu, Chao Wang, Haowei He, and
  Menglin Yang.
\newblock Where does long-context supervision actually go? effective-context
  exposure balancing.
\newblock \emph{arXiv preprint arXiv:2605.10544}, 2026.

\bibitem[Zhu et~al.(2025)Zhu, Chen, Wang, Yu, Zhao, and
  Jia]{zhu_2025_generalization}
Mingkang Zhu, Xi~Chen, Zhongdao Wang, Bei Yu, Hengshuang Zhao, and Jiaya Jia.
\newblock Enhancing {LLM} knowledge learning through generalization.
\newblock \emph{arXiv preprint arXiv:2503.03705}, 2025.

\bibitem[Zwaan and Radvansky(1998)]{zwaan_radvansky_1998}
Rolf~A. Zwaan and Gabriel~A. Radvansky.
\newblock Situation models in language comprehension and memory.
\newblock \emph{Psychological Bulletin}, 123\penalty0 (2):\penalty0 162--185,
  1998.
\newblock Cited as motivation, not as evidence.

\end{thebibliography}

\end{document}